\documentclass[sigconf,authordraft]{acmart}
\AtBeginDocument{%
  \providecommand\BibTeX{{%
    \normalfont B\kern-0.5em{\scshape i\kern-0.25em b}\kern-0.8em\TeX}}}

\copyrightyear{2020}
\acmYear{2020}
\setcopyright{acmcopyright}\acmConference[KDD '20]{Proceedings of the 26th ACM SIGKDD Conference on Knowledge Discovery and Data Mining}{August 23--27, 2020}{Virtual Event, CA, USA}
\acmBooktitle{Proceedings of the 26th ACM SIGKDD Conference on Knowledge Discovery and Data Mining (KDD '20), August 23--27, 2020, Virtual Event, CA, USA}
\acmPrice{15.00}
\acmDOI{10.1145/3394486.3403049}
\acmISBN{978-1-4503-7998-4/20/08}

\usepackage{subfig}
\usepackage{amsmath,enumitem}
\usepackage[ruled,vlined,linesnumbered,noresetcount]{algorithm2e}
\usepackage{threeparttable,color}
\usepackage{booktabs}
\usepackage{multirow}

\DeclareMathOperator*{\argmin}{arg\,min}

\begin{document}
\fancyhead{}
\title{Graph Structure Learning for Robust Graph Neural Networks}


\author{Wei Jin}
\affiliation{%
  \institution{Michigan State University}
}
\email{jinwei2@msu.edu}
\author{Yao Ma}
\affiliation{%
  \institution{Michigan State University}
}
\email{mayao4@msu.edu}
\author{Xiaorui Liu}
\affiliation{%
  \institution{Michigan State University}
}
\email{xiaorui@msu.edu}
\author{Xianfeng Tang}
\affiliation{%
  \institution{The Pennsylvania State University}
}
\email{tangxianfeng@outlook.com}
\author{Suhang Wang}
\affiliation{%
  \institution{The Pennsylvania State University}
}
\email{szw494@psu.edu}
\author{Jiliang Tang}
\affiliation{%
  \institution{ Michigan State University}
}
\email{tangjili@msu.edu}



\newcommand{\modelname}{Pro-GNN }
\newcommand{\SVDpaper}{GCN-SVD }
\newcommand{\metattack}{\textit{metattack }}
\newcommand{\nettack}{\textit{\textit{nettack }}}

\begin{abstract}
Graph Neural Networks (GNNs) are powerful tools in representation learning for graphs. However, recent studies show that GNNs are vulnerable to carefully-crafted perturbations, called adversarial attacks. Adversarial attacks can easily fool GNNs in making predictions for downstream tasks. The vulnerability to adversarial attacks has raised increasing concerns for applying GNNs in safety-critical applications. Therefore, developing robust algorithms to defend adversarial attacks is of great significance. A natural idea to defend adversarial attacks is to clean the perturbed graph. It is evident that real-world graphs share some intrinsic properties. For example, many real-world graphs are low-rank and sparse, and the features of two adjacent nodes tend to be similar. In fact, we find that adversarial attacks are likely to violate these graph properties. Therefore, in this paper, we explore these properties to defend adversarial attacks on graphs.  In particular, we propose a general framework Pro-GNN, which can jointly learn a structural graph and a robust graph neural network model from the perturbed graph guided by these properties. Extensive experiments on real-world graphs demonstrate that the proposed framework achieves significantly better performance compared with the state-of-the-art defense methods, even when the graph is heavily perturbed.
We release the implementation of Pro-GNN to our DeepRobust repository for adversarial attacks and defenses~\footnote{\url{https://github.com/DSE-MSU/DeepRobust}}. The specific experimental settings to reproduce our results can be found in https://github.com/ChandlerBang/Pro-GNN.

\end{abstract}

\maketitle

\section{Introduction}

Graphs are ubiquitous data structures in numerous domains, such as chemistry (molecules), finance (trading networks) and social media (the Facebook friend network). With their prevalence, it is particularly important to learn effective representations of graphs and then apply them to solve downstream tasks. Recent years have witnessed great success from Graph Neural  Networks (GNNs) \cite{li2015gated,graphsage,kipf2016semi,gat} in representation learning of graphs. GNNs follow a message-passing scheme \cite{mpnn}, where the node embedding is obtained by aggregating and transforming the embeddings of its neighbors. Due to the good performance, GNNs have been applied to various analytical tasks including node classification \cite{kipf2016semi}, link prediction \cite{vgae}, and recommender systems \cite{pinsage}. 

Although promising results have been achieved, recent studies have shown that GNNs are vulnerable to adversarial attacks \cite{jin2020adversarial,nettack,mettack,rl-s2v,preprocess}. In other words, the performance of GNNs can greatly degrade under an unnoticeable perturbation in graphs. The lack of robustness of these models can lead to severe consequences for critical applications pertaining to the safety and privacy. For example, in credit card fraud detection, fraudsters can create several transactions with only a few high-credit users to disguise themselves, thus escaping from the detection based on GNNs. Hence, developing robust GNN models to resist adversarial attacks is of significant importance. Modifying graph data can perturb either node features or graph structures. However, given the complexity of structural information, the majority of existing adversarial attacks on graph data have focused on modifying graph structure especially adding/deleting/rewiring edges~\cite{xu2019adversarial}. Thus, in this work, we aim to defend against the most common setting of adversarial attacks on graph data, i.e., poisoning adversarial attacks on graph structure. Under this setting, the graph structure has already been perturbed by modifying edges before training GNNs while node features are not changed.

\begin{figure*}[ht]%
\vskip -0.2em
    \centering
    \subfloat[Singular Values\label{fig:singular-value}]
        {{\includegraphics[width=0.25\linewidth]{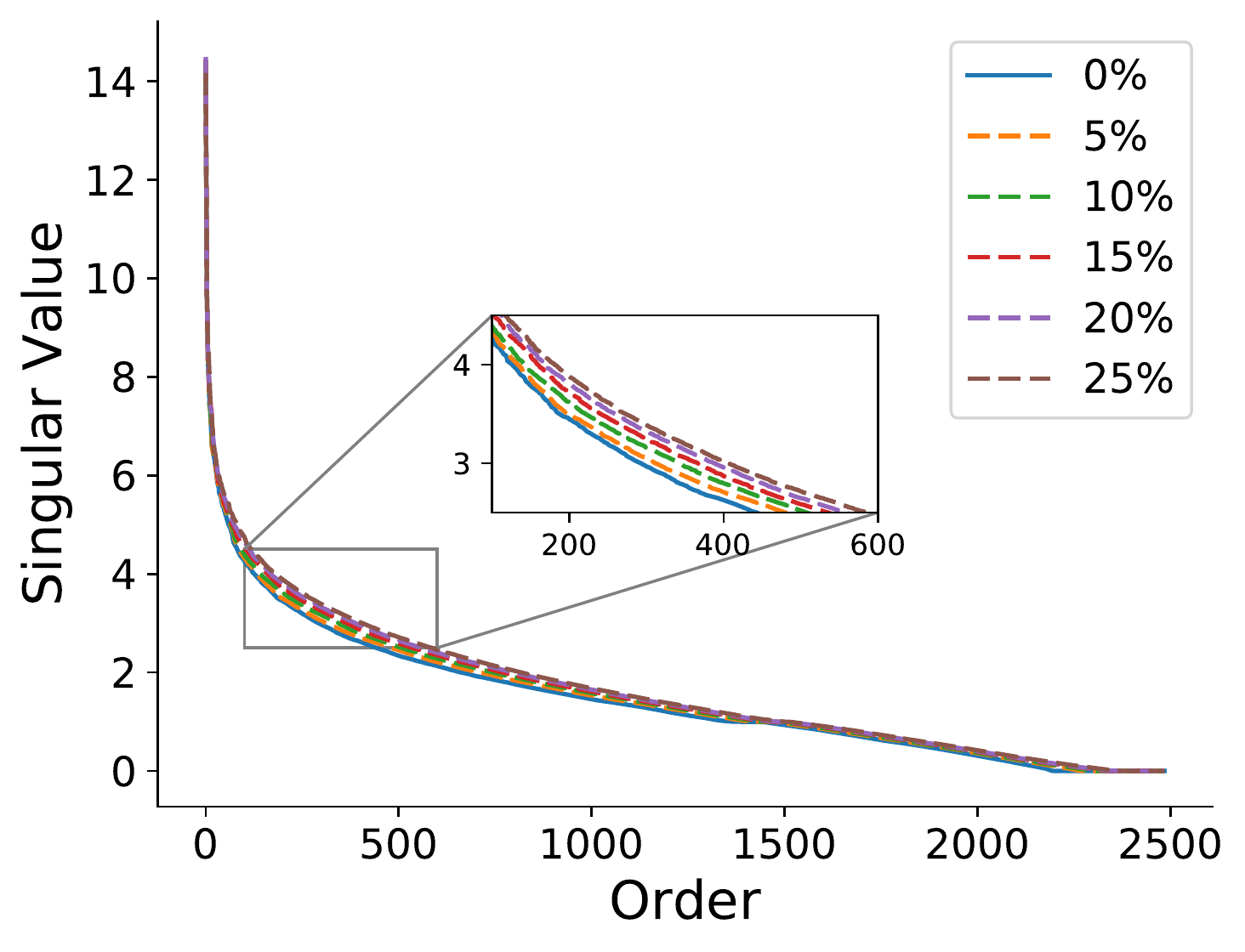}}}%
    \subfloat[Rank Growth\label{fig:rank-growth}]{{\includegraphics[width=0.25\linewidth]{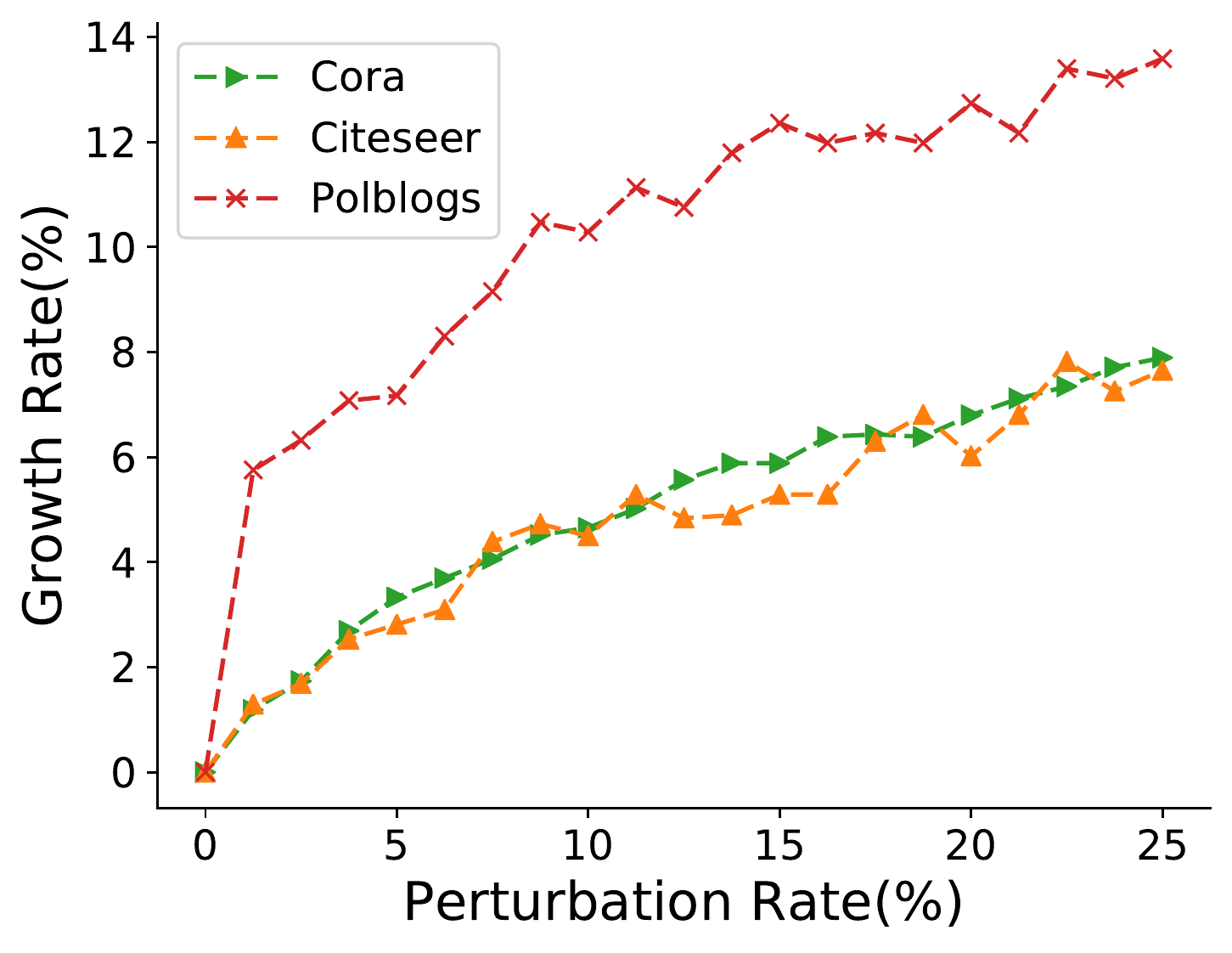} }}%
    \subfloat[Rank Decrease Rate \label{fig:remove-edge}]{{\includegraphics[width=0.25\linewidth]{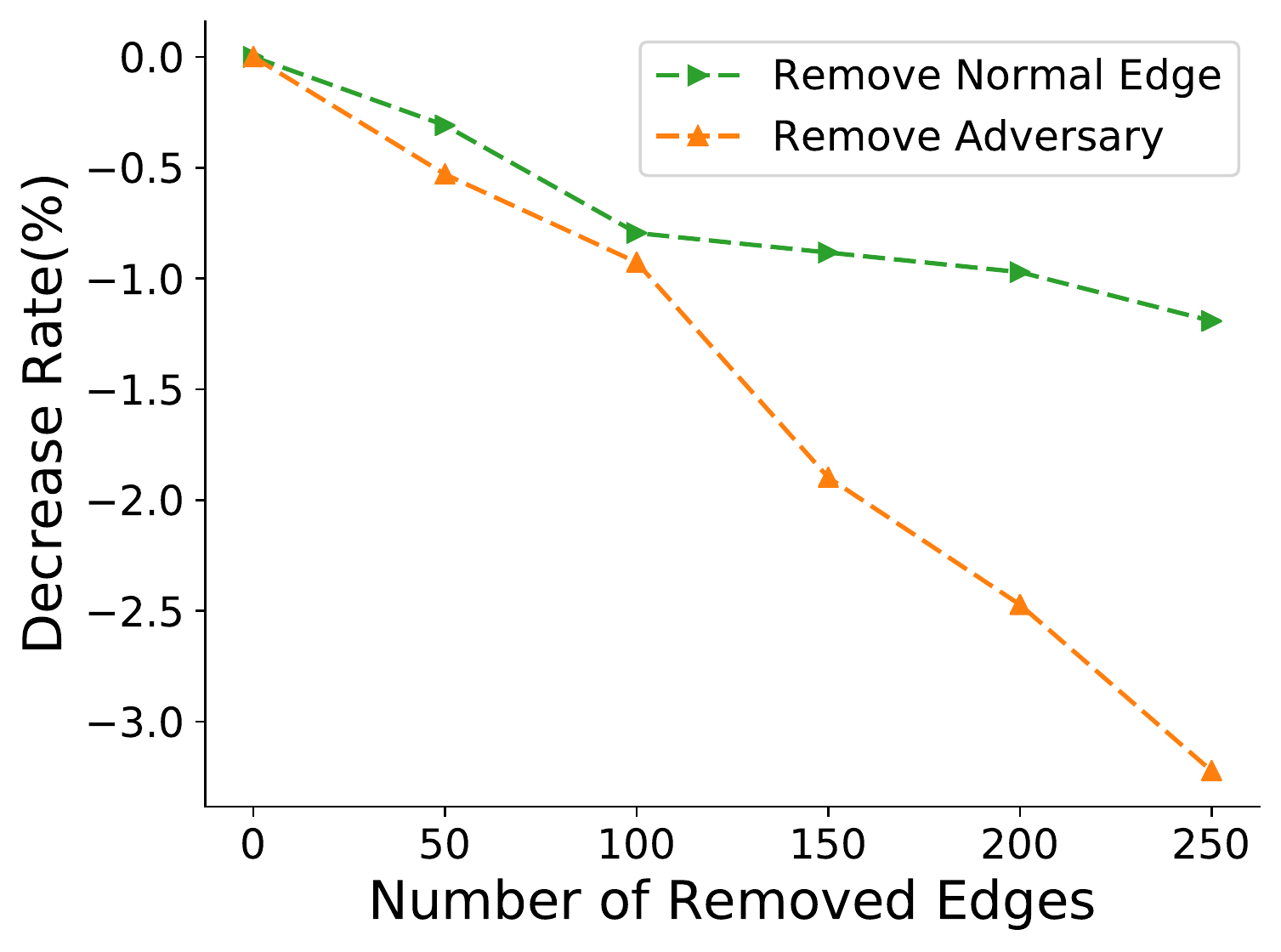}}}%
    \subfloat[Feature Smoothness\label{fig:feature-smooth}]{{\includegraphics[width=0.25\linewidth]{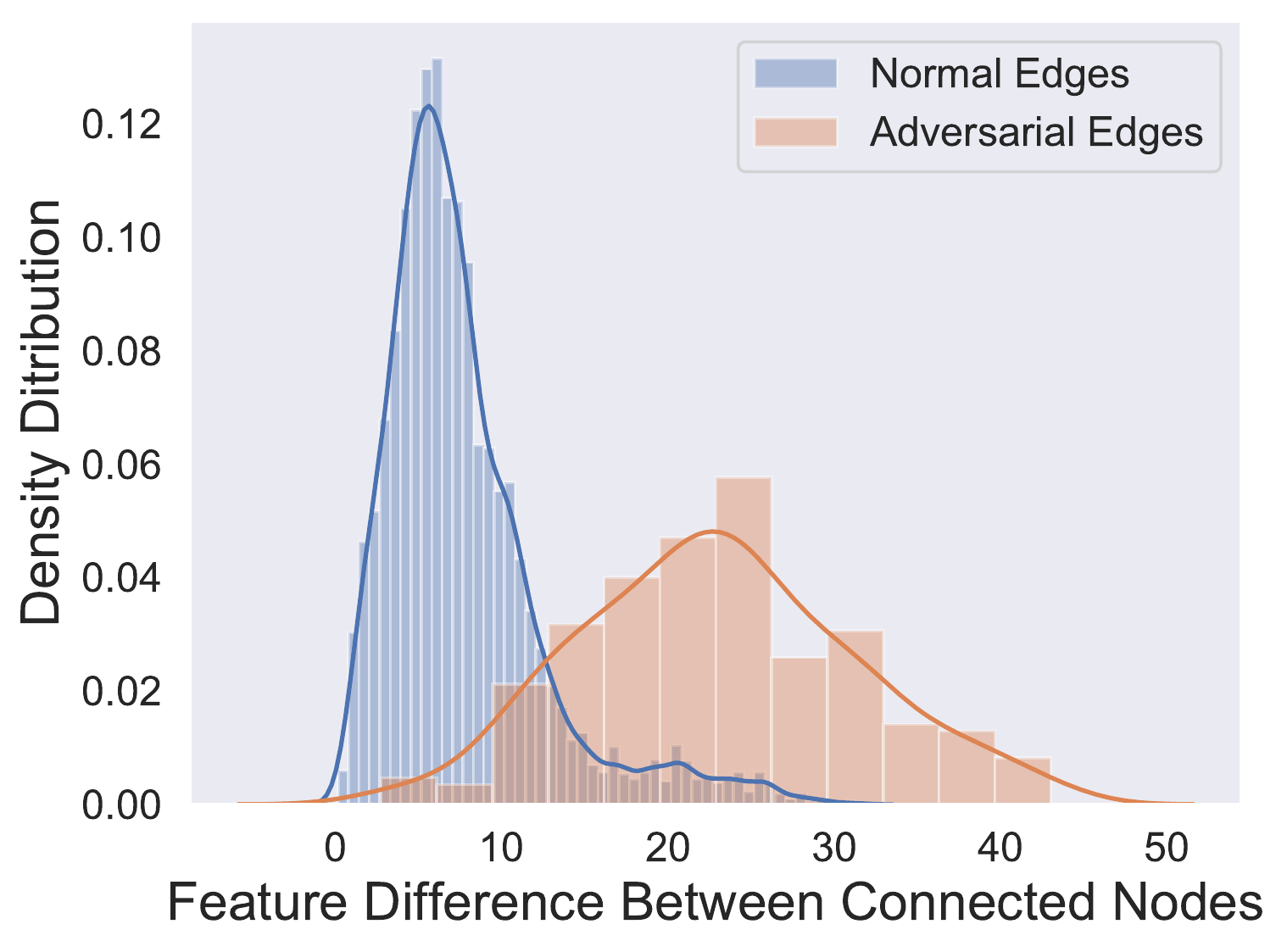} }}%
    \qquad
    \vskip -1.2em
    \caption{An illustrative example on the property changes of the adjacency matrix by adversarial attacks}%
    \label{fig:observation}
    \vskip -1em
\end{figure*}

One perspective to design an effective defense algorithm is to clean the perturbed graph such as removing the adversarial edges and restoring the deleted edges~\cite{rgcn,pagnn}. The key challenge from this perspective is what criteria we should follow to clean the perturbed graph. It is well known that real-world graphs often share certain properties. First, many real-world clean graphs are low-rank and sparse \cite{kezhou_lesong2013learning}. For instance, in a social network, most individuals are connected with only a small number of neighbors and there are only a few factors influencing the connections among users~\cite{kezhou_lesong2013learning,fortunato2010community}. Second, connected nodes in a clean graph are likely to share similar features or attributes (or feature smoothness)~\cite{mcpherson2001birds}. For example, in a citation network, two connected publications often share similar topics~\cite{kipf2016semi}. Figure~\ref{fig:observation} demonstrates these properties of clean and poisoned graphs. Specifically, we apply the state-of-the-art graph poisoning attack, \metattack\cite{mettack}, to perturb the graph data and visualize the graph properties before and after \textit{mettack}. As shown in Figure~\ref{fig:singular-value}, \metattack enlarges the singular values of the adjacency matrix and Figure~\ref{fig:rank-growth} illustrates that \metattack quickly increases the rank of adjacency matrix. Moreover, when we remove the adversarial and normal edges from the perturbed graph respectively, we observe that removing adversarial edges reduces the rank faster than removing normal edges as demonstrated in Figure~\ref{fig:remove-edge}. In addition, we depict the density distribution of feature difference of connected nodes of the attacked graph in Figure~\ref{fig:feature-smooth}. It is observed that \metattack tends to connect nodes with large feature difference. Observations from Figure~\ref{fig:observation} indicate that adversarial attacks could violate these properties.  Thus, these properties have the potential to serve as the guidance to clean the perturbed graph. However, work of exploring these properties to build robust graph neural networks is rather limited.




In this paper, we target on exploring graph properties of sparsity, low rank and feature smoothness to design robust graph neural networks. Note that there could be more properties to be explored and we would like to leave it as future work. In essence, we are faced with two challenges: (i) how to learn clean graph structure from poisoned graph data guided by these properties; and (ii) how to jointly learn parameters for robust graph neural network and the clean structure. To solve these two challenges, we propose a general framework \underline{Pro}perty \underline{GNN} (Pro-GNN) to simultaneously learn the clean graph structure from perturbed graph and  GNN parameters to defend against  adversarial attacks. Extensive experiments on a variety of real-world graphs demonstrate that our proposed model can effectively defend against different types of adversarial attacks and outperforms the state-of-the-art defense methods.


The rest of the paper is organized as follows.  In Section $2$, we review some of the related work. In Section $3$, we introduce notations and formally define the problem. We explain our proposed framework in Section $4$ and report our experimental results in Section $5$. Finally, we conclude the work with future directions in Section $6$.

\section{Related Work}

In line with the focus of our work, we briefly describe related work on GNNs, and adversarial attacks and defense for graph data.

\subsection{Graph Neural Networks}
Over the past few years, graph neural networks have achieved great success in solving machine learning problems on graph data. To learn effective representation of graph data, two main families of GNNs have been proposed, i.e., spectral methods and spatial methods. The first family learns node representation based on graph spectral theory~\cite{kipf2016semi, bruna2013spectral, ChebNet}. Bruna et al.~\cite{bruna2013spectral} generalize the convolution operation from Euclidean data to non-Euclidean data by using the Fourier basis of a given graph. To simplify spectral GNNs,  Defferrard et al.~\cite{ChebNet} propose ChebNet and utilize Chebyshev polynomials as the convolution filter. Kipf et al.~\cite{kipf2016semi} propose GCN and simplify ChebNet by using its first-order approximation. Further, Simple Graph Convolution (SGC)~\cite{wu2019simplifying} reduces the graph convolution to a linear model but still achieves competitive performance. The second family of models define graph convolutions in the spatial domain as aggregating and transforming local information~\cite{graphsage, mpnn, gat}. For instance, DCNN~\cite{atwood2016diffusion} treats graph convolutions as a diffusion process and assigns a certain transition probability for information transferred from one node to the adjacent node. Hamilton et al.~\cite{graphsage} propose to learn aggregators by sampling and aggregating neighbor information.  Veli{\v{c}}kovi{\'c} et al.~\cite{gat} propose graph attention network (GAT) to learn different attention scores for neighbors when aggregating information. To further improve the training efficiency, FastGCN~\cite{chen2018fastgcn} interprets graph convolutions as integral transforms of embedding functions under probability measures and performs importance sampling to sample a fixed number of nodes for each layer. For a thorough review, we please refer the reader to recent surveys~ \cite{zhou2018graph-survey, wu2019comprehensive-survey, battaglia2018relational-survey}.

\subsection{Adversarial Attacks and Defense for GNNs}
Extensive studies have demonstrated that deep learning models are vulnerable to adversarial attacks. In other words, slight or unnoticeable perturbations to the input can fool a neural network to output a wrong prediction. GNNs also suffer this problem~\cite{jin2020adversarial, nettack, rl-s2v, mettack, rewiring2019attack, liu2019unified, bojchevski2019adversarial, preprocess}. Different from image data, the graph structure is discrete and the nodes are dependent of each other, thus making it far more challenging. 
The \nettack~\cite{nettack} generates unnoticeable perturbations by preserving degree distribution and imposing constraints on feature co-occurrence. RL-S2V~\cite{rl-s2v} employs reinforcement learning to generate adversarial attacks. However, both of the two methods are designed for targeted attack and can only degrade the performance of GNN on target nodes. To perturb the graph globally, \metattack~\cite{mettack} is proposed to generate poisoning attacks based on meta-learning. Although increasing efforts have been devoted to developing adversarial attacks on graph data, the research about improving the robustness of GNNs has just started recently~\cite{rgcn, preprocess, pagnn, zugner2019certifiable}. One way to solve the problem is to learn a robust network by penalizing the attention scores of adversarial edges. RGCN~\cite{rgcn} is to model Gaussian distributions as hidden layers to absorb the effects of adversarial attacks in the variances. 
PA-GNN~\cite{pagnn} leverages supervision knowledge from clean graphs and applies a meta-optimization way to learn attention scores for robust graph neural networks. However, it requires additional graph data from similar domain. The other way is to preprocess the perturbed graphs to get clean graphs and train GNNs on the clean ones. Wu et. al \cite{preprocess} have found that attackers tend to connect to nodes with different features and they propose to remove the links between dissimilar nodes. Entezari et al. \cite{all-you-need-is-low-rank} have observed that \nettack results in changes in high-rank spectrum of the graph and propose to preprocess the graph with its low-rank approximations. However, due to the simplicity of two-stage preprocessing methods, they may fail to counteract complex global attacks. 

Different from the aforementioned defense methods, we aim to explore important graph properties to recover the clean graph while learning the GNN parameters simultaneously, which enables the proposed model to extract intrinsic structure from perturbed graph under different attacks.

\section{Problem Statement}
Before we present the problem statement, we first introduce some notations and basic concepts. The Frobenius norm of a matrix ${\bf S}$ is defined by $||{\bf S}||^2_F=\Sigma_{ij}{ {\bf S}^2_{ij}}$. The $\ell_1$ norm of a matrix ${\bf S}$ is given by $||{\bf S}||_1=\Sigma_{ij}{|{\bf S}_{ij}|}$ and the nuclear norm of a matrix ${\bf S}$ is defined as $||{\bf S}||_* = \Sigma^{rank({\bf S})}_{i=1}{\sigma_i}$ , where $\sigma_i$ is the $i$-th singular value of ${\bf S}$. $({\bf S})_+$ denotes the element-wise positive part of matrix ${\bf S}$ where ${\bf S}_{ij}=\max\{{\bf S}_{ij}, 0\}$ and $sgn( {\bf S})$ indicates the sign matrix of $\bf S$ where $sgn( {\bf S})_{ij} = 1, 0,$ or $-1$ if ${\bf S}_{ij}$ >0, =0, or <0, respectively. We use $\odot$ to denote Hadamard product of matrices. Finally, we use $tr({\bf S})$ to indicate the trace of matrix ${\bf S}$, i.e., $tr({\bf S}) = \sum_{i} \mathbf{S}_{ii}$.  

Let $\mathcal{G}=(\mathcal{V},\mathcal{E})$ be a graph, where $\mathcal{V}$ is the set of $N$ nodes $\{v_1, v_2, ..., v_N\}$ and $\mathcal{E}$ is the set of edges. The edges describe the relations between nodes and can also be represented by an adjacency matrix $\mathbf{A} \in \mathbb{R}^{N \times N}$ where $\mathbf{A}_{ij}$ denotes the relation between nodes $v_i$ and $v_j$. Furthermore, we use ${\bf X} = [{\bf x}_1,{\bf x}_2,\ldots,{\bf x}_N] \in\mathbb{R}^{{N}\times{d}}$ to denote the node feature matrix where ${\bf x}_i$ is the feature vector of the node $v_i$. Thus a graph can also be denoted as $\mathcal{G}=({\bf A},{\bf X})$. 
Following the common node classification setting, only a part of nodes $\mathcal{V}_{L}=\{v_1, v_2, ..., v_l\}$ are associated with corresponding labels $\mathcal{Y}_L =\{y_1, y_2, ..., y_l\}$ where $y_i$ denotes the label of $v_i$.

Given a graph $\mathcal{G}=({\bf A}, {\bf X})$ and the partial labels $\mathcal{Y}_L$, the goal of node classification for GNN is to learn a function $f_\theta: \mathcal{V}_L\rightarrow\mathcal{Y}_L$ that maps the nodes to the set of labels so that $f_\theta$ can predict labels of unlabeled nodes. The objective function can be formulated as  
\begin{equation}
\min _{\theta} \mathcal{L}_{GNN}( {\theta}, {\bf A}, {\bf X}, \mathcal{Y}_L)=\sum_{v_i \in \mathcal{V}_{L}} \ell\left(f_{\theta}({\bf X}, {\bf A})_{i}, y_{i}\right),
\end{equation}
where $\theta$ is the parameters of $f_\theta$, $f_\theta( {\bf X}, {\bf A})_i$ is the prediction of node $v_i$ and $\ell(\cdot, \cdot)$ is to measure the difference between prediction and true label such as cross entropy.
Though there exist a number of different GNN methods, in this work, we focus on Graph Convolution Network (GCN) in \cite{kipf2016semi}. Note that it is straightforward to extend the proposed framework to other GNN models. Specifically, a two-layer GCN with $\theta=({\bf W}_1, {\bf W}_2)$ implements $f_\theta$ as 
\begin{equation}
f_{\theta}( {\bf X}, {\bf A})=\operatorname{softmax}\left(\hat{\bf A}\thinspace \operatorname{\sigma} \left(\hat{\bf A}\thinspace {\bf X}\thinspace {\bf W}_{1}\right) {\bf W}_{2}\right),
\end{equation}
where $\hat{\bf A}=\tilde{\bf D}^{-1 / 2}({\bf A}+{\bf I}) \tilde{\bf D}^{-1 / 2}$ and $\tilde{\bf D}$ is the diagonal matrix of ${\bf A}+ {\bf I}$ with $\tilde{\bf D}_{ii} = 1 + \sum_{j} {\bf A}_{ij}$. $\operatorname{\sigma}$ is the activation function such as ReLU. 

With aforementioned notations and definitions, the problem we aim to study in this work can be formally stated as: 
\vskip 0.5em
\textit{Given $\mathcal{G}=({\bf A, X})$ and partial node label $\mathcal{V}_L$ with ${\bf A}$ being poisoned by adversarial edges and feature matrix ${\bf X}$ unperturbed, simultaneously learn a clean graph structure with the graph adjacency matrix ${\bf S}\in\mathcal{S}=[0,1]^{N{\times}N}$ and the GNN parameters ${\theta}$ to improve node classification performance for unlabeled nodes.}


\section{The Proposed Framework}

\begin{figure}[t]
    \centering
    \includegraphics[width=\linewidth]{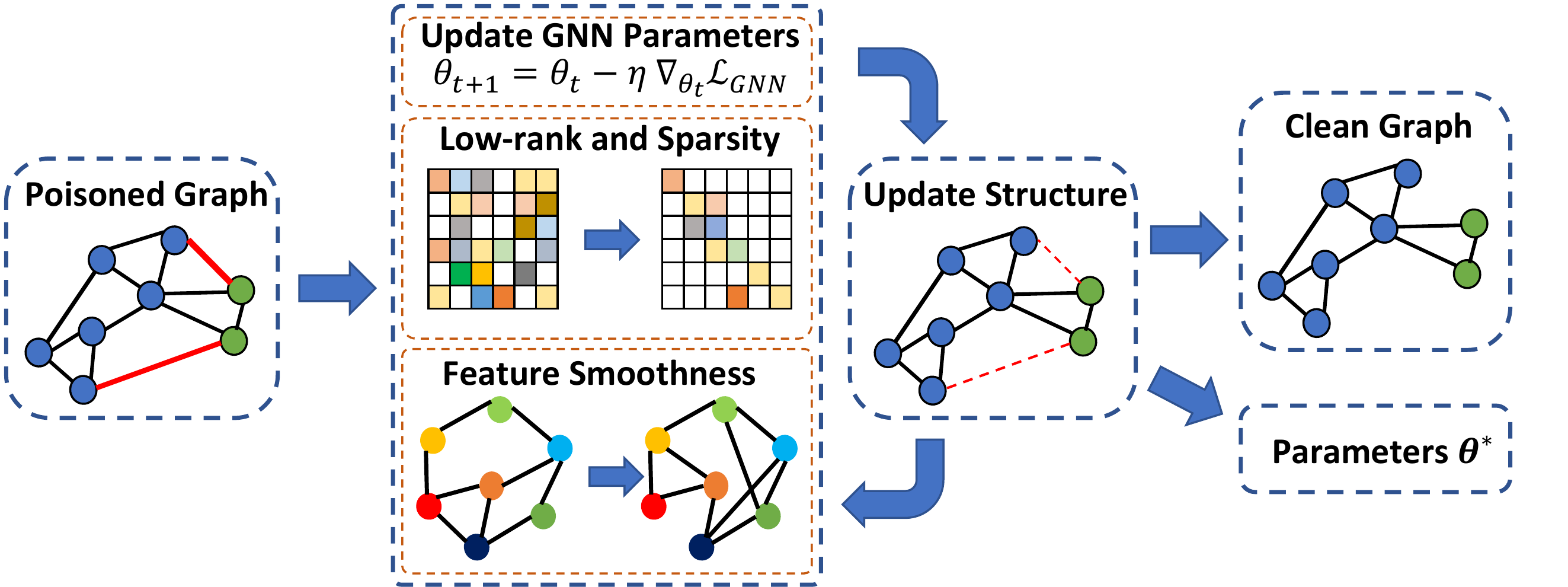}
    \vskip -1em
    \caption{Overall framework of Pro-GNN. Dash lines indicate smaller weights.}
    \label{fig:framework}
    \vskip -1em
\end{figure}

Adversarial attacks generate carefully-crafted perturbation on graph data. We refer to the carefully-crafted perturbation as adversarial structure.  Adversarial structure can cause the performance of GNNs to drop rapidly. Thus, to defend adversarial attacks, one natural strategy is to eliminate the crafted adversarial structure, while maintaining the intrinsic graph structure. In this work, we aim to achieve the goal by exploring graph structure properties of low rank, sparsity and feature smoothness. The illustration of the framework is shown in Figure~\ref{fig:framework}, where edges in black are normal edges and edges in red are adversarial edges introduced by an attacker to reduce the node classification performance. To defend against the attacks, Pro-GNN iteratively reconstructs the clean graph by preserving the low rank, sparsity, and feature smoothness properties of a graph so as to reduce the negative effects of adversarial structure. Meanwhile, to make sure that the reconstructed graph can help node classification, Pro-GNN simultaneously updates the GNN parameters on the reconstructed graph by solving the optimization problem in an alternating schema. 
In the following subsections, we will give the details of the proposed framework.  
 

\subsection{Exploring Low rank and Sparsity Properties}

Many real-world graphs are naturally low-rank and sparse as the entities usually tend to form communities and would only be connected with a small number of neighbors \cite{kezhou_lesong2013learning}. Adversarial attacks on GCNs tend to add adversarial edges that link nodes of different communities as this is more efficient to reduce node classification performance of GCN. Introducing links connecting nodes of different communities in a sparse graph can significantly increase the rank of the adjacency matrix and enlarge the singular values, thus damaging the low rank and sparsity properties of graphs, which is verified in  Figure~\ref{fig:singular-value} and Figure~\ref{fig:rank-growth}. 
Thus, to recover the clean graph structure from the noisy and perturbed graph, one potential way is to learn a clean adjacency matrix $\mathbf{S}$ close to the adjacency matrix of the poisoned graph by enforcing the new adjacency matrix with the properties of low rank and sparsity. As demonstrated in Figure\ref{fig:remove-edge}, the rank decreases much faster by removing adversarial edges than by removing normal edges. This implies that the low rank and sparsity constraint can remove the adversarial edges instead of normal edges.  Given the adjacency matrix ${\bf A}$ of a poisoned graph, we can formulate the above process as a  structure learning problem~\cite{richard2012estimation,jiang2019semi}: 
\begin{equation}
    \argmin_{ {\bf S}\in\mathcal{S}} \mathcal{L}_0 = \| {\bf A} - {\bf S} \|^2_F + R( {\bf S} ), ~~ s.t., {\bf S} = {\bf S}^\top.
 \label{eq:reg}
\end{equation}
Since adversarial attacks target on performing unnoticeable perturbations to graphs, the first term $\| {\bf A} - {\bf S} \|^2_F$ ensures that the new adjacency matrix ${\bf S}$ should be close to ${\bf A}$. As we assume that the graph are undirected, the new adjacency matrix should be symmetric, i.e., ${\bf S} = {\bf S}^\top$.  $R( {\bf S} )$ denotes the constraints on ${\bf S}$ to enforce the properties of low rank and sparsity. According to \cite{candes2009exact-exact-matrix-completion-convex-optimization, koltchinskii2011nuclear-for-noisy, richard2012estimation}, minimizing the $\ell_{1}$ norm and the nuclear norm of a matrix can force the matrix to be sparse and low-rank, respectively.  Hence, to ensure a sparse and low-rank  graph, we want to minimize the $\ell_{1}$ norm and the nuclear norm of ${\bf S}$.  Eq.~(\ref{eq:reg}) can be rewritten as: 
\begin{equation}
    \argmin_{ {\bf S}\in\mathcal{S}} \mathcal{L}_0 = \| {\bf A} - {\bf S}\|^2_F + \alpha\| {\bf S}\|_1 + \beta\|{\bf S}\|_*, ~~ s.t., {\bf S} = {\bf S}^\top
    \label{eq:low-ranksparse},
\end{equation}
where $\alpha$ and $\beta$ are predefined parameters that control the contributions of the properties of sparsity and low rank, respectively. One important benefit to minimize the nuclear norm $\| {\bf S} \|_*$ is that we can reduce every singular value, thus alleviating the impact of enlarging singular values from adversarial attacks.


\subsection{Exploring Feature Smoothness}
It is evident that connected nodes in a graph are likely to share similar features. In fact, this observation has been made on graphs from numerous domains. For example, two connected users in a social graph are likely to share similar attributes~\cite{mcpherson2001birds}, two linked web pages in the webpage graph tend to have similar contents~\cite{websites-similar} and two connected papers in the citation network usually have similar topics~\cite{kipf2016semi}. Meanwhile, recently it is demonstrated that adversarial attacks on graphs tend to connect nodes with distinct features~\cite{preprocess}. Thus, we aim to ensure the feature smoothness in the learned graph. The feature smoothness can be captured by the following term $\mathcal{L}_{s}$:
\begin{equation}
    \mathcal{L}_{s} =\frac{1}{2}\sum_{i,j=1}^N{ {\bf S}_{ij}( {\bf x}_i- {\bf x}_j)^2}
    \label{eq:xLap},
\end{equation}
\noindent where ${\bf S}$ is the new adjacency matrix, ${\bf S}_{ij}$ indicates the connection of $v_i$ and $v_j$ in the learned graph and $({\bf x}_i- {\bf x}_j)^2$ measures the feature difference between $v_i$ and $v_j$. $\mathcal{L}_{s}$ can be rewritten as: 
\begin{equation}
    \mathcal{L}_{s} = tr({\bf X}^\top {\bf L} {\bf X}),
    \label{eq:xLap}
\end{equation}
where ${\bf L} = {\bf D}- {\bf S}$ is the graph Laplacian matrix of ${\bf S}$ and ${\bf D}$ is the diagonal matrix of ${\bf S}$. In this work, we use normalized Laplacian matrix $\hat{\bf L}={\bf D}^{-1/2} {\bf L} {\bf D}^{-1/2}$ instead of ${\bf L}$ to make feature smoothness independent on the degrees of the graph nodes \cite{ando2007learning}, i.e., 
\begin{equation}
    \mathcal{L}_{s} = tr({\bf X}^T \hat{\bf L} {\bf X}) = \frac{1}{2}\sum_{i,j=1}^N{{\bf S}_{ij}(\frac{ {\bf x}_{i}}{\sqrt{d_i}}-\frac{{\bf x}_{j}}{\sqrt{d_j}})^2},
\end{equation} 
\noindent where $d_i$ denotes the degree of $v_i$ in the learned graph. In the learned graph, if $v_i$ and $v_j$ are connected (i.e., ${\bf S}_{ij} \neq 0$), we expect that the feature difference $({\bf x}_i- {\bf x}_j)^2$ should be small. In other words, if the features between two connected node are quite different, $\mathcal{L}_s$ would be very large. Therefore, the smaller $\mathcal{L}_s$ is, the smoother features ${\bf X}$ are on the graph ${\bf S}$. Thus, to fulfill the feature smoothness in the learned graph, we should minimize $\mathcal{L}_s$. Therefore, we can add the feature smoothness term to the objective function of Eq.~(\ref{eq:low-ranksparse}) to penalize rapid changes in features between adjacent nodes as: 
\begin{equation}
    \argmin_{ {\bf S}\in\mathcal{S}} \mathcal{L} = \mathcal{L}_0 + \lambda\cdot \mathcal{L}_{s} = \mathcal{L}_0 + \lambda\thinspace{tr({\bf X}^T \hat{\bf L} {\bf X})},~s.t.,~ {\bf S} = {\bf S}^\top,
    \label{eq:nonGCN}
\end{equation}
\noindent where $\lambda$ is a predefined parameter to control the contribution from feature smoothness.

\subsection{Objective Function of Pro-GNN}
Intuitively, we can follow the preprocessing strategy~\cite{preprocess,all-you-need-is-low-rank} to defend against adversarial attacks -- we first learn a graph from the poisoned graph via Eq.~(\ref{eq:nonGCN}) and then train a GNN model based on the learned graph. However, with such a two-stage strategy, the learned graph may be suboptimal for the GNN model on the given task. Thus, we propose a better strategy to jointly learn the graph structure and the GNN model for a specific downstream task. We empirically show that jointly learning GNN model and the adjacency matrix is better than two stage one in Sec~\ref{sec:two_stage_one_stage}.  The final objective function of Pro-GNN is given as
\begin{align}
\small
\label{eq:obj}
       & \argmin_{{\bf S}\in\mathcal{S}, {\bf \theta}} \mathcal{L} = \mathcal{L}_0 + \lambda \mathcal{L}_s + \gamma \mathcal{L}_{GNN}  \\
    &     = \| {\bf A} - {\bf S}\|^2_F + \alpha\| {\bf S}\|_1 + \beta \| {\bf S}\|_* + \gamma \mathcal{L}_{GNN}({\bf \theta}, {\bf S}, {\bf X}, \mathcal{Y}_L) + \lambda tr( {\bf X}^T\hat{\bf L} {\bf X}) \nonumber \\
    &\quad s.t. \quad \quad {\bf S} = {\bf S}^\top \nonumber,
    \label{p1}
\end{align} 
where $\mathcal{L}_{GNN}$ is a loss function for the GNN model that is controlled by a predefined parameter $\gamma$. Another benefit of this formulation is that the information from $\mathcal{L}_{GNN}$ can also guide the graph learning process to defend against adversarial attacks since the goal of graph adversarial attacks is to maximize $\mathcal{L}_{GNN}$.

\subsection{An Optimization Algorithm}
\label{sec:opt}
Jointly optimizing $\theta$ and ${\bf S}$ in Eq.(\ref{eq:obj}) is challenging. The constraints on ${\bf S}$ further exacerbate the difficulty. Thus, in this work, we use an alternating optimization schema to iteratively update $\theta$ and ${\bf S}$. 
\vskip 0.5em
\noindent{}\textbf{Update $\boldsymbol{ \theta}$.} To update  $\theta$, we fix $\mathbf{S}$ and remove terms that are irrelevant to $\theta$, then the objective function in Eq.(\ref{eq:obj}) reduces to:
\begin{equation}
\min _{\theta} \mathcal{L}_{GNN}( {\theta}, {\bf S}, {\bf X}, \mathcal{Y}_L)=\sum_{u \in \mathcal{V}_{L}} \ell\left(f_{\theta}({\bf X}, {\bf S})_{u}, y_{u}\right),
\end{equation}
which is a typical GNN optimization problem and we can learn ${\bf \theta}$ via stochastic gradient descent.
\vskip 0.5em
\noindent{}\textbf{Update $\mathbf{S}$.}
Similarly, to update $\mathbf{S}$, we fix $\theta$ and arrive at
\begin{equation}
\begin{aligned}
    \min_{\mathbf{S}} \mathcal{L}({\bf S, A}) + \alpha\| {\bf S}\|_1 + \beta \| {\bf S}\|_*  \quad s.t.,  \quad {\bf S} = {\bf S}^\top, {\bf S} \in \mathcal{S},
\end{aligned}
\end{equation}
where $\mathcal{L}({\bf S, A})$ is defined as
\begin{equation}
    \mathcal{L}({\bf S, A})=\| {\bf A} - {\bf S}\|^2_F + \mathcal{L}_{GNN}({\bf \theta}, {\bf S}, {\bf X}, Y) + \lambda tr( {\bf X}^T\hat{\bf L} {\bf X}).
\end{equation}
Note that both $\ell_{1}$ norm and nuclear norm are non-differentiable. 
For optimization problem with only one non-diffiential regularizer $R(S)$, we can use Forward-Backward splitting methods \cite{combettes2011proximal}. The idea is to alternate a gradient descent step and a proximal step as:  
\begin{equation}
{\mathbf{S}^{(k)}} {=\operatorname{prox}_{\eta R}\left(\mathbf{S}^{(k-1)}-\eta\nabla_S{\mathcal{L}(S, A)}\right)},
\end{equation}
where $\eta$ is the learning rate, $\operatorname{prox}_{R}$ is the proximal operator as: 
\begin{equation}
    \operatorname{prox}_{R}({\bf Z})=\argmin_{{\bf S}\in\mathbb{R}^{N\times{N}}}{\frac{1}{2}\| {\bf S}- {\bf Z}||^2_F+R{(\bf S)}}.
\end{equation}
In particular, the proximal operator of $\ell_{1}$ norm and nuclear norm can be represented as \cite{richard2012estimation, shrinkage-thresholding},
\begin{equation}
    \operatorname{prox}_{\alpha||.||_1}( {\bf Z})=sgn({\bf Z})\odot (|{\bf Z}|-\alpha)_+,
\end{equation}
\begin{equation}
    \operatorname{prox}_{\beta||.||_*}({\bf Z})= {\bf U} \thinspace diag((\sigma_i-\beta)_+)_{i} {\bf V}^T,
\end{equation}
where $ {\bf Z} = {\bf U} \thinspace diag(\sigma_1, ..., \sigma_n){\bf V}^\top$ is the singular value decomposition of ${\bf Z}$. To optimize objective function with two non-differentiable regularizers, Richard et al. \cite{generalized-forward-backward-splitting} introduce the Incremental Proximal Descent method based on the introduced proximal operators. By iterating the updating process in a cyclic manner, we can update ${\bf S}$ as follows,
\begin{equation}
\centering 
\left\{
\begin{array}{ll}
{\mathbf{S}^{(k)}} \thinspace{} {=\mathbf{S}^{(k-1)}-\eta \cdot \nabla_{\mathbf{S}}
\left(\mathcal{L}(\mathbf{S}, \mathbf{A})\right)}, \\ 
{\mathbf{S}^{(k)}} \thinspace{} {=\operatorname{prox}_{\eta \beta\|\cdot\|_{*}}\left(\mathbf{S}^{(k)}\right)}, \\ {\mathbf{S}^{(k)}}  \thinspace{} {=\operatorname{prox}_{\eta \alpha\|\cdot\|_{1}}\left(\mathbf{S}^{(k)}\right)}.\end{array}\right.
\end{equation}

After we learn a relaxed ${\bf S}$, we project ${\bf S}$ to satisfy the constraints.  For the symmetric constraint, we let ${\bf S} = \frac{ {\bf S} + {\bf S}^\top}{2}$. For the constraint ${\bf S}_{ij} \in [0,1]$, we project ${\bf S}_{ij} <0$ to $0$ and  ${\bf S}_{ij} > 1$ to $1$.  We denote these projection operations as $ P_\mathcal{S}({\bf S})$.   
\vskip 0.5em
\noindent{}\textbf{Training Algorithm.} With these updating and projection rules, the optimization algorithm is shown in Algorithm~\ref{alg:1}. In line 1, we first initialize the estimated graph ${\bf S}$ as the poisoned graph ${\bf A}$. In line 2, we randomly initialize the GNN parameters. From lines 3 to 10, we update ${\bf S}$ and the GNN parameters ${\bf \theta}$ alternatively and iteratively. Specifically, we train the GNN parameters in each iteration while training the graph reconstruction model every $\tau$ iterations. 
\begin{algorithm}[t]
  \KwData{Adjacency matrix ${\bf A}$, Attribute matrix ${\bf X}$, Labels $\mathcal{Y}_L$, Hyper-parameters ${\alpha, \beta, \gamma, \lambda, \tau}$, Learning rate ${\eta,\eta'}$}
  \KwResult{Learned adjacency ${\bf S}$, GNN parameters $\theta$}
  Initialize ${\bf S}\leftarrow{{\bf A}}$\\
  Randomly initialize $\theta$\\
  \While{Stopping condition is not met}{
    $ {\bf S} \leftarrow{{\bf S}-\eta \nabla_S(\| {\bf S}- {\bf A}\|_F^2+\gamma\mathcal{L}_{GNN}+\lambda\mathcal{L}_s)}$\\
    ${\bf S}\leftarrow{\operatorname{prox}_{\eta\beta||.||_*}({\bf S})}$\\
    ${\bf S}\leftarrow{\operatorname{prox}_{\eta\alpha||.||_1}({\bf S})}$\\
    ${\bf S}\leftarrow{P_\mathcal{S}({\bf S})}$\\
    \For{i=1 to $\tau$}{
      $g\leftarrow{\frac{\partial \mathcal{L}_{GNN}(\theta, {\bf S}, {\bf X}, {\mathcal{Y}_L})}{\partial \theta}}$\\
      $ {\bf \theta}\leftarrow{\bf \theta} - \eta' g$
    }
  }
  \textbf{Return} ${\bf S}, {\bf \theta}$
  \caption{\modelname}
  \label{alg:1}
\end{algorithm}

\section{Experiments}
In this section, we evaluate the effectiveness of Pro-GNN against different graph adversarial attacks. In particular,  we aim to answer the following questions:
\begin{itemize}[leftmargin=*]
\item{\textbf{RQ1}} How does Pro-GNN perform compared to the state-of-the-art defense methods under different adversarial attacks?
\item{\textbf{RQ2}} Does the learned graph work as expected? 
\item{\textbf{RQ3}} How do different properties affect performance of Pro-GNN.
\end{itemize}
Before presenting our experimental results and observations, we first introduce the experimental settings.

\subsection{Experimental settings}
\subsubsection{Datasets}
Following~\cite{nettack,mettack}, 
we validate the proposed approach on four benchmark datasets, including three citation graphs, i.e., Cora, Citeseer and Pubmed, and one blog graph, i.e., Polblogs. The statistics of the datasets are shown in Table~\ref{dataset}. Note that in the Polblogs graph, node features are not available.  In this case, we set the attribute matrix to  $N\times N$ identity matrix. 

\begin{table}[tb]
\vskip -1em
\centering
\caption{Dataset Statistics. Following \cite{nettack, mettack, all-you-need-is-low-rank}, we only consider the largest connected component (LCC).}
\vskip -1em
\begin{tabular}{c|cccc}
\toprule
         & \textbf{$\mathbf{N_{LCC}}$} & \textbf{$\mathbf{E_{LCC}}$}  & \textbf{Classes} & \textbf{Features} \\ \midrule
Cora     & 2,485 & 5,069  & 7       & 1,433     \\ 
Citeseer & 2,110 & 3,668  & 6       & 3,703     \\ 
Polblogs & 1,222 & 16,714 & 2       & /        \\ 
Pubmed	 &19,717 & 44,338 & 3       & 500 \\ \bottomrule
\end{tabular}
\vspace{-1em}
\label{dataset}
\end{table}

\begin{table*}[t] 
\small
\caption{Node classification performance (Accuracy$\pm$Std) under non-targeted attack (\metattack).}
\vskip -1em
\label{table:metattack}
\begin{threeparttable}
\begin{tabular}{c|c|ccccccc}
\toprule
Dataset        & Ptb Rate (\%)  & GCN                     & GAT                     & RGCN           & GCN-Jaccard\footnote{}    & GCN-SVD        & Pro-GNN-fs              & Pro-GNN\footnote{}                 \\ \midrule
\multirow{6}{*}{Cora}     & 0                                                         & 83.50$\pm$0.44          & \textbf{83.97$\pm$0.65} & 83.09$\pm$0.44 & 82.05$\pm$0.51 & 80.63$\pm$0.45 & 83.42$\pm$0.52          & 82.98$\pm$0.23          \\
                          & 5                                                         & 76.55$\pm$0.79          & 80.44$\pm$0.74          & 77.42$\pm$0.39 & 79.13$\pm$0.59 & 78.39$\pm$0.54 & \textbf{82.78$\pm$0.39} & 82.27$\pm$0.45          \\
                          & 10                                                        & 70.39$\pm$1.28          & 75.61$\pm$0.59          & 72.22$\pm$0.38 & 75.16$\pm$0.76 & 71.47$\pm$0.83 & 77.91$\pm$0.86          & \textbf{79.03$\pm$0.59} \\
                          & 15                                                        & 65.10$\pm$0.71          & 69.78$\pm$1.28          & 66.82$\pm$0.39 & 71.03$\pm$0.64 & 66.69$\pm$1.18 & 76.01$\pm$1.12          & \textbf{76.40$\pm$1.27} \\
                          & 20                                                        & 59.56$\pm$2.72          & 59.94$\pm$0.92          & 59.27$\pm$0.37 & 65.71$\pm$0.89 & 58.94$\pm$1.13 & 68.78$\pm$5.84          & \textbf{73.32$\pm$1.56} \\
                          & 25                                                        & 47.53$\pm$1.96          & 54.78$\pm$0.74          & 50.51$\pm$0.78 & 60.82$\pm$1.08 & 52.06$\pm$1.19 & 56.54$\pm$2.58          & \textbf{69.72$\pm$1.69} \\
\midrule
\multirow{6}{*}{Citeseer} & 0                                                         & 71.96$\pm$0.55          & 73.26$\pm$0.83          & 71.20$\pm$0.83 & 72.10$\pm$0.63 & 70.65$\pm$0.32 & 73.26$\pm$0.38          & \textbf{73.28$\pm$0.69} \\
                          & 5                                                         & 70.88$\pm$0.62          & 72.89$\pm$0.83          & 70.50$\pm$0.43 & 70.51$\pm$0.97 & 68.84$\pm$0.72 & \textbf{73.09$\pm$0.34} & 72.93$\pm$0.57          \\
                          & 10                                                        & 67.55$\pm$0.89          & 70.63$\pm$0.48          & 67.71$\pm$0.30 & 69.54$\pm$0.56 & 68.87$\pm$0.62 & 72.43$\pm$0.52          & \textbf{72.51$\pm$0.75} \\
                          & 15                                                        & 64.52$\pm$1.11          & 69.02$\pm$1.09          & 65.69$\pm$0.37 & 65.95$\pm$0.94 & 63.26$\pm$0.96 & 70.82$\pm$0.87          & \textbf{72.03$\pm$1.11} \\
                          & 20                                                        & 62.03$\pm$3.49          & 61.04$\pm$1.52          & 62.49$\pm$1.22 & 59.30$\pm$1.40 & 58.55$\pm$1.09 & 66.19$\pm$2.38          & \textbf{70.02$\pm$2.28} \\
                          & 25                                                        & 56.94$\pm$2.09          & 61.85$\pm$1.12          & 55.35$\pm$0.66 & 59.89$\pm$1.47 & 57.18$\pm$1.87 & 66.40$\pm$2.57          & \textbf{68.95$\pm$2.78} \\
\midrule
\multirow{6}{*}{Polblogs} & 0                                                         & \textbf{95.69$\pm$0.38} & 95.35$\pm$0.20          & 95.22$\pm$0.14 & -              & 95.31$\pm$0.18 & 93.20$\pm$0.64          & -                       \\
                          & 5                                                         & 73.07$\pm$0.80          & 83.69$\pm$1.45          & 74.34$\pm$0.19 & -              & 89.09$\pm$0.22 & \textbf{93.29$\pm$0.18} & -                       \\
                          & 10                                                        & 70.72$\pm$1.13          & 76.32$\pm$0.85          & 71.04$\pm$0.34 & -              & 81.24$\pm$0.49 & \textbf{89.42$\pm$1.09} & -                       \\
                          & 15                                                        & 64.96$\pm$1.91          & 68.80$\pm$1.14          & 67.28$\pm$0.38 & -              & 68.10$\pm$3.73 & \textbf{86.04$\pm$2.21} & -                       \\
                          & 20                                                        & 51.27$\pm$1.23          & 51.50$\pm$1.63          & 59.89$\pm$0.34 & -              & 57.33$\pm$3.15 & \textbf{79.56$\pm$5.68} & -                       \\
                          & 25                                                        & 49.23$\pm$1.36          & 51.19$\pm$1.49          & 56.02$\pm$0.56 & -              & 48.66$\pm$9.93 & \textbf{63.18$\pm$4.40} & -                       \\
\midrule
\multirow{6}{*}{Pubmed}   & 0                                                         & 87.19$\pm$0.09          & 83.73$\pm$0.40          & 86.16$\pm$0.18 & 87.06$\pm$0.06 & 83.44$\pm$0.21 & \textbf{87.33$\pm$0.18} & 87.26$\pm$0.23          \\
                          & 5                                                         & 83.09$\pm$0.13          & 78.00$\pm$0.44          & 81.08$\pm$0.20 & 86.39$\pm$0.06 & 83.41$\pm$0.15 & \textbf{87.25$\pm$0.09} & 87.23$\pm$0.13          \\
                          & 10                                                        & 81.21$\pm$0.09          & 74.93$\pm$0.38          & 77.51$\pm$0.27 & 85.70$\pm$0.07 & 83.27$\pm$0.21 & \textbf{87.25$\pm$0.09} & 87.21$\pm$0.13          \\
                          & 15                                                        & 78.66$\pm$0.12          & 71.13$\pm$0.51          & 73.91$\pm$0.25 & 84.76$\pm$0.08 & 83.10$\pm$0.18 & \textbf{87.20$\pm$0.09} & 87.20$\pm$0.15          \\
                          & 20                                                        & 77.35$\pm$0.19          & 68.21$\pm$0.96          & 71.18$\pm$0.31 & 83.88$\pm$0.05 & 83.01$\pm$0.22 & 87.09$\pm$0.10          & \textbf{87.15$\pm$0.15} \\
                          & 25                                                        & 75.50$\pm$0.17          & 65.41$\pm$0.77          & 67.95$\pm$0.15 & 83.66$\pm$0.06 & 82.72$\pm$0.18 & 86.71$\pm$0.09          & \textbf{86.76$\pm$0.19} \\ 
\bottomrule
\end{tabular}
\begin{tablenotes}
   \item[1 2] JaccardGCN and Pro-GNN cannot be directly applied to datasets where node features are not available.
  \end{tablenotes}
\end{threeparttable}

\end{table*}

\subsubsection{Baselines}
To evaluate the effectiveness of Pro-GNN, we compare it with the state-of-the-art GNN and defense models by using the adversarial attack repository DeepRobust~\cite{li2020deeprobust}:
\begin{itemize}[leftmargin=*]
    \item \textbf{GCN} \cite{kipf2016semi}: while there exist a number of different Graph Convolutional Networks (GCN) models, we focus on the most representative one \cite{kipf2016semi}.  
    \item \textbf{GAT} \cite{gat}: Graph Attention Netowork (GAT) is composed of attention layers which can learn different weights to different nodes in the neighborhood. It is often used as a baseline to defend against adversarial attacks. 
    \item \textbf{RGCN} \cite{rgcn}: RGCN models node representations as gaussian distributions to absorb effects of adversarial attacks. It also employs attention mechanism to penalize nodes with high variance.
    \item \textbf{GCN-Jaccard} \cite{preprocess}: Since attackers tend to connect nodes with dissimilar features or different labels, GCN-Jaccard preprocesses the network by eliminating edges that connect nodes with jaccard similarity of features smaller than threshold $\tau$. Note that this method only works when node features are available.
    \item \textbf{\SVDpaper} \cite{all-you-need-is-low-rank}: This is another preprocessing method to resist adversarial attacks. It is noted that \nettack is a high-rank attack, thus \SVDpaper  proposes to vaccinate GCN with the low-rank approximation of the perturbed graph. Note that it originally targets at defending against \textit{nettack}, however, it is straightforward to extend it to non-targeted and random attacks.
\end{itemize}
In addition to representative baselines, we also include one variant of the proposed framework, Pro-GNN-fs, which is the variant by eliminating the feature smoothness term (or setting $\lambda = 0$). 

\subsubsection{Parameter Settings}

For each graph, we randomly choose 10\% of nodes for training, 10\% of nodes for validation and the remaining 80\% of nodes for testing. For each experiment, we report the average performance of 10 runs.  The hyper-parameters of all the models are tuned based on the loss and accuracy on validation set. For GCN and GAT, we adopt the default parameter setting in the author's implementation. For RGCN, the number of hidden units are tuned from $\{16, 32, 64, 128\}$. For GCN-Jaccard, the threshold of similarity for removing dissimilar edges is chosen from $\{0.01, 0.02, 0.03, 0.04, 0.05, 0.1\}$. For \SVDpaper, the reduced rank of the perturbed graph is tuned from $\{5, 10, 15, 50, 100, 200\}$. 

\subsection{Defense Performance}
To answer the first question, we evaluate the node classification performance of Pro-GNN against three types of attacks, i.e., non-targeted attack, targeted attack and random attack:


\begin{itemize}[leftmargin=*]
    \item \textbf{Targeted Attack:} Targeted attack generates attacks on specific nodes and aims to fool GNNs on these target nodes. We adopt \nettack \cite{nettack} for the targeted attack method, which is the state-of-the-art targeted attack on graph data. 
    \item \textbf{Non-targeted Attack:} Different from targeted attack, the goal of non-targeted attack is to degrade the overall performance of GNNs on the whole graph. We adopt one representative non-targeted attack, \metattack \cite{mettack}  .
    \item \textbf{Random Attack:} It randomly injects fake edges into the graph. It can also be viewed as adding random noise to the clean graph. 
\end{itemize}

\begin{figure*}[t]%
    \centering
    \subfloat[Cora]{{\includegraphics[width=0.25\linewidth]{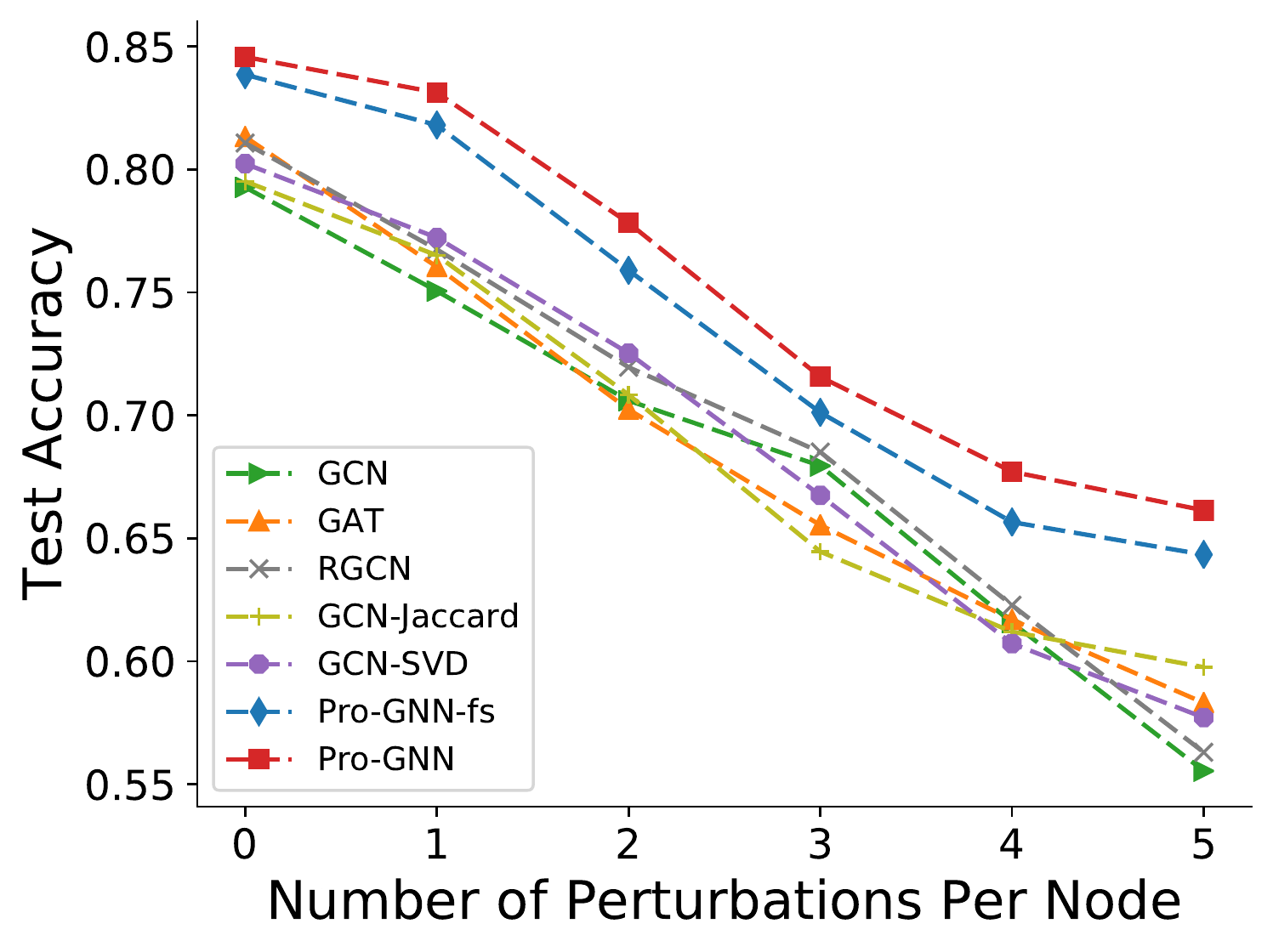} }}%
    \subfloat[Citeseer]{{\includegraphics[width=0.25\linewidth]{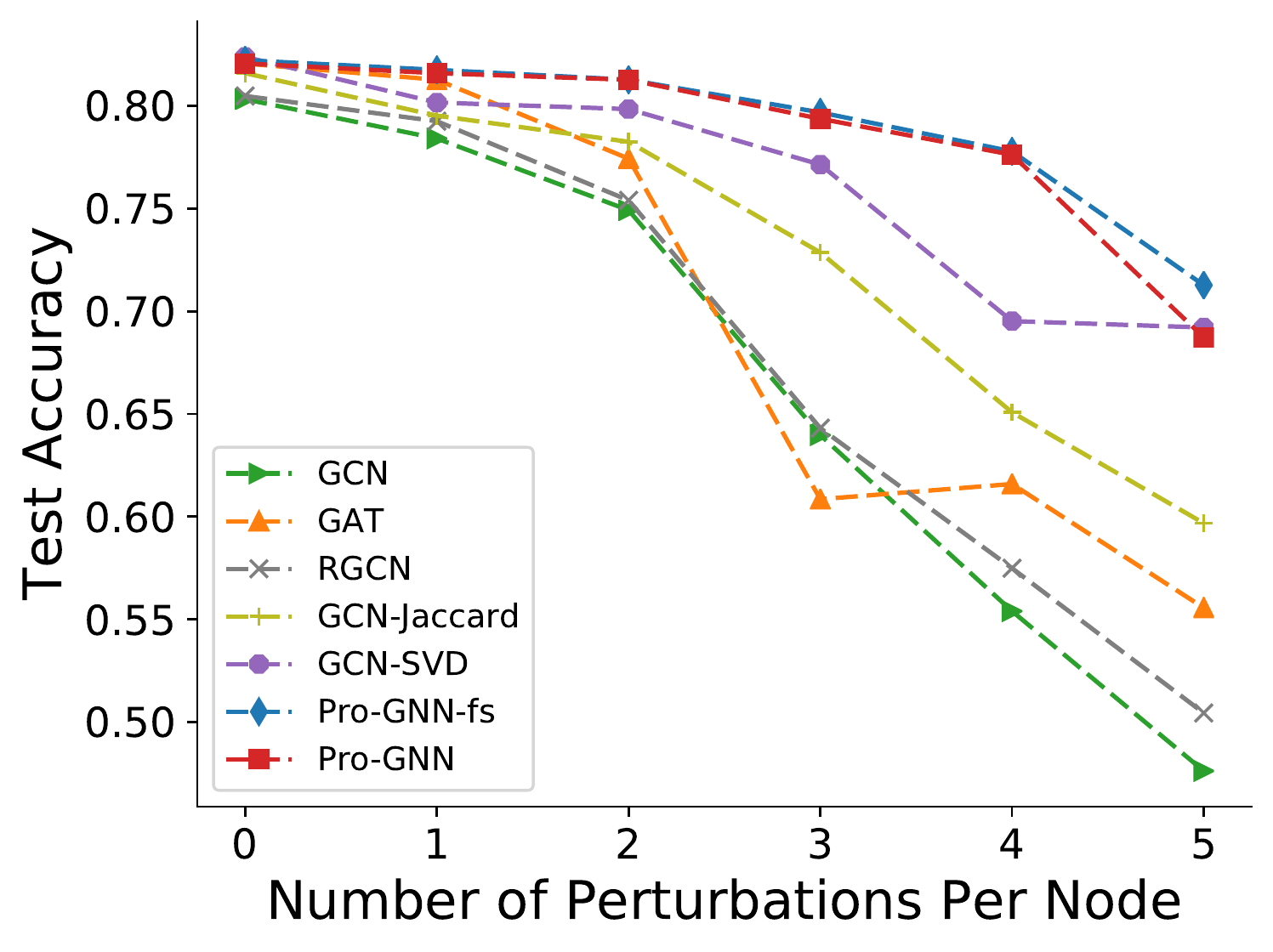} }}%
    \subfloat[Polblogs]{{\includegraphics[width=0.25\linewidth]{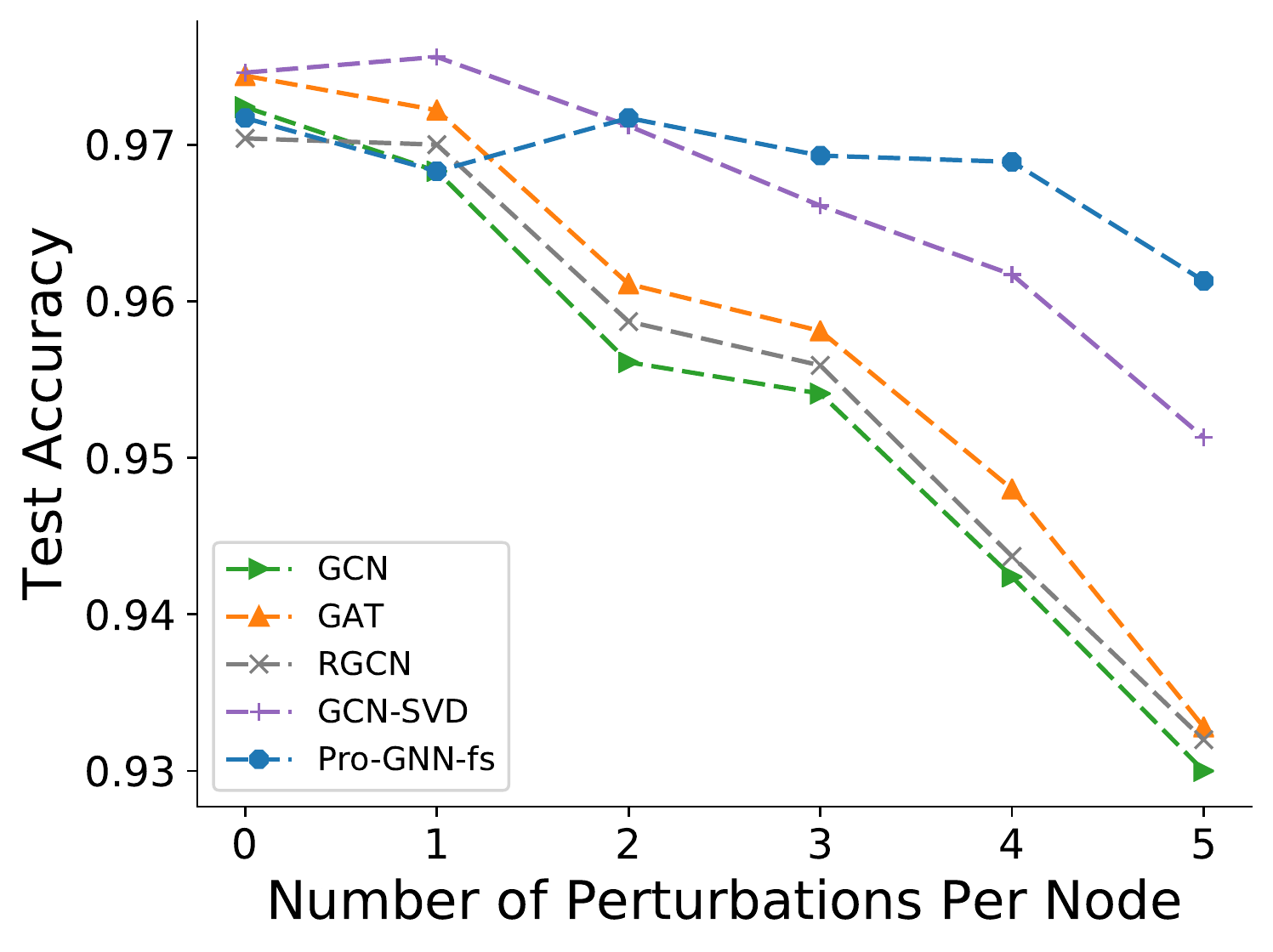} }}%
    \subfloat[Pubmed]{{\includegraphics[width=0.25\linewidth]{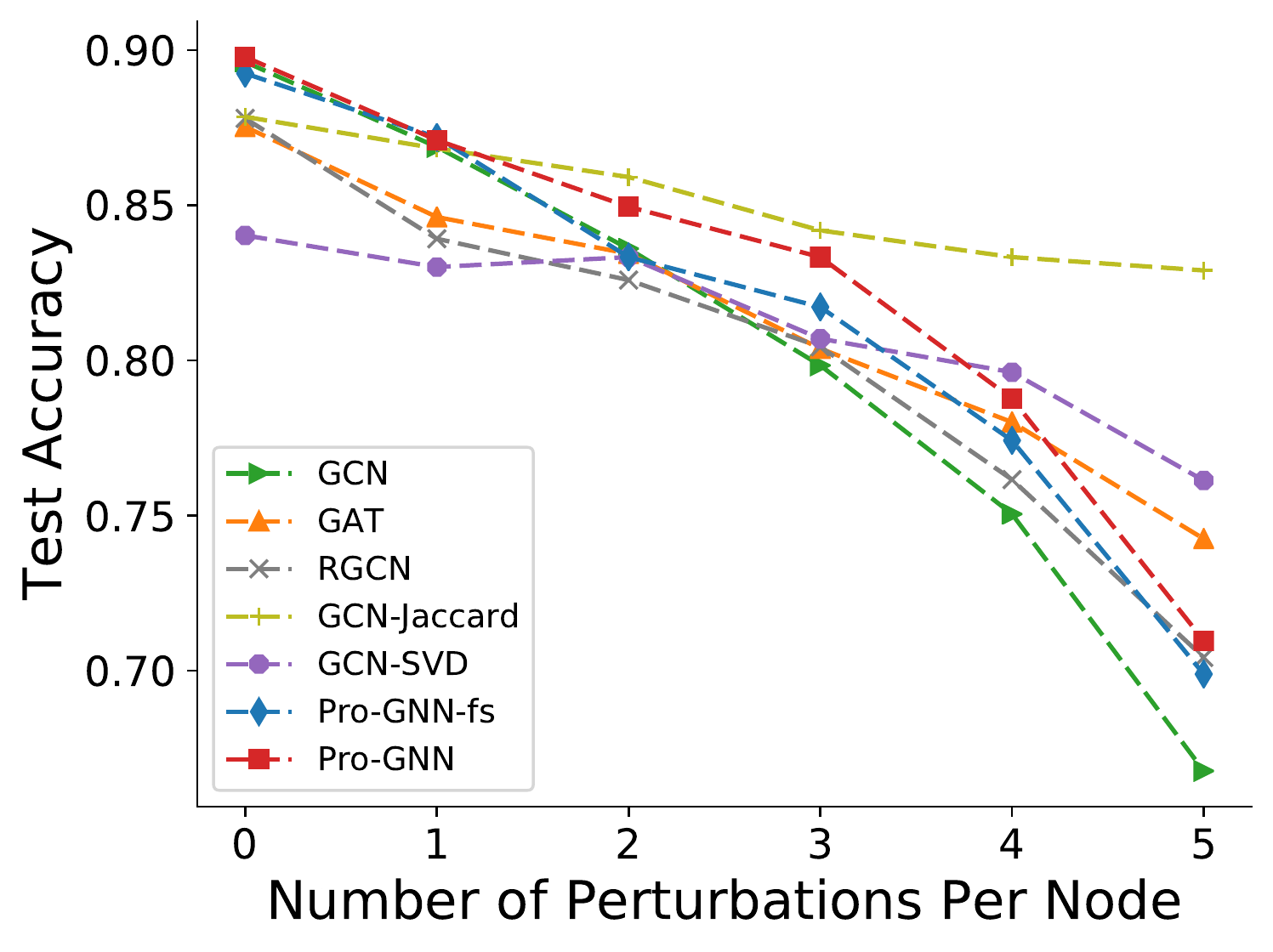} }}%
    \vskip -1.2em
    \caption{Results of different models under \nettack}%
    \vskip -1em
    \label{fig:nettack}%
\end{figure*}
\begin{figure*}[htb]%
\vskip -1em
    \subfloat[Cora]{{\includegraphics[width=0.25\linewidth]{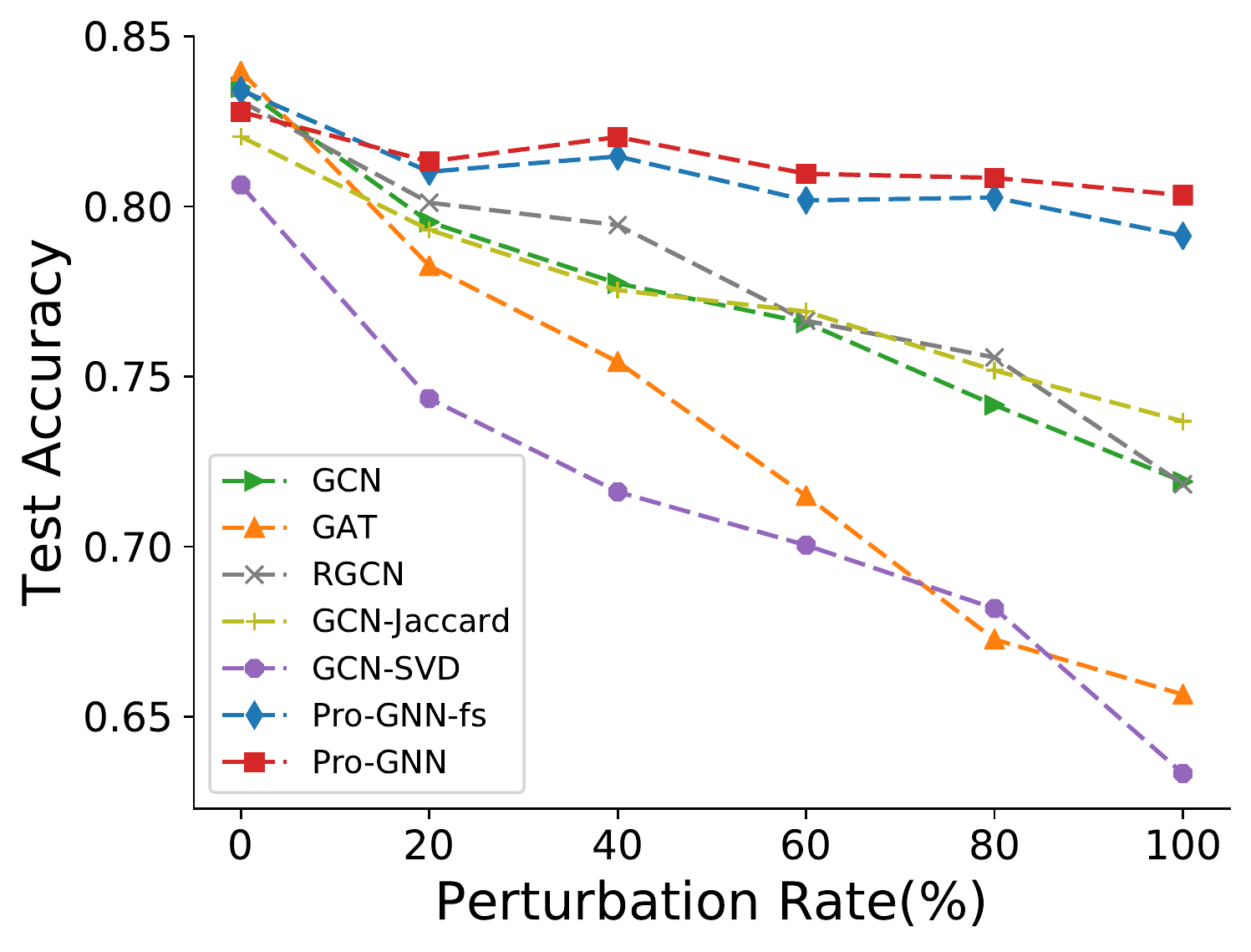} }}%
    \subfloat[Citeseer]{{\includegraphics[width=0.25\linewidth]{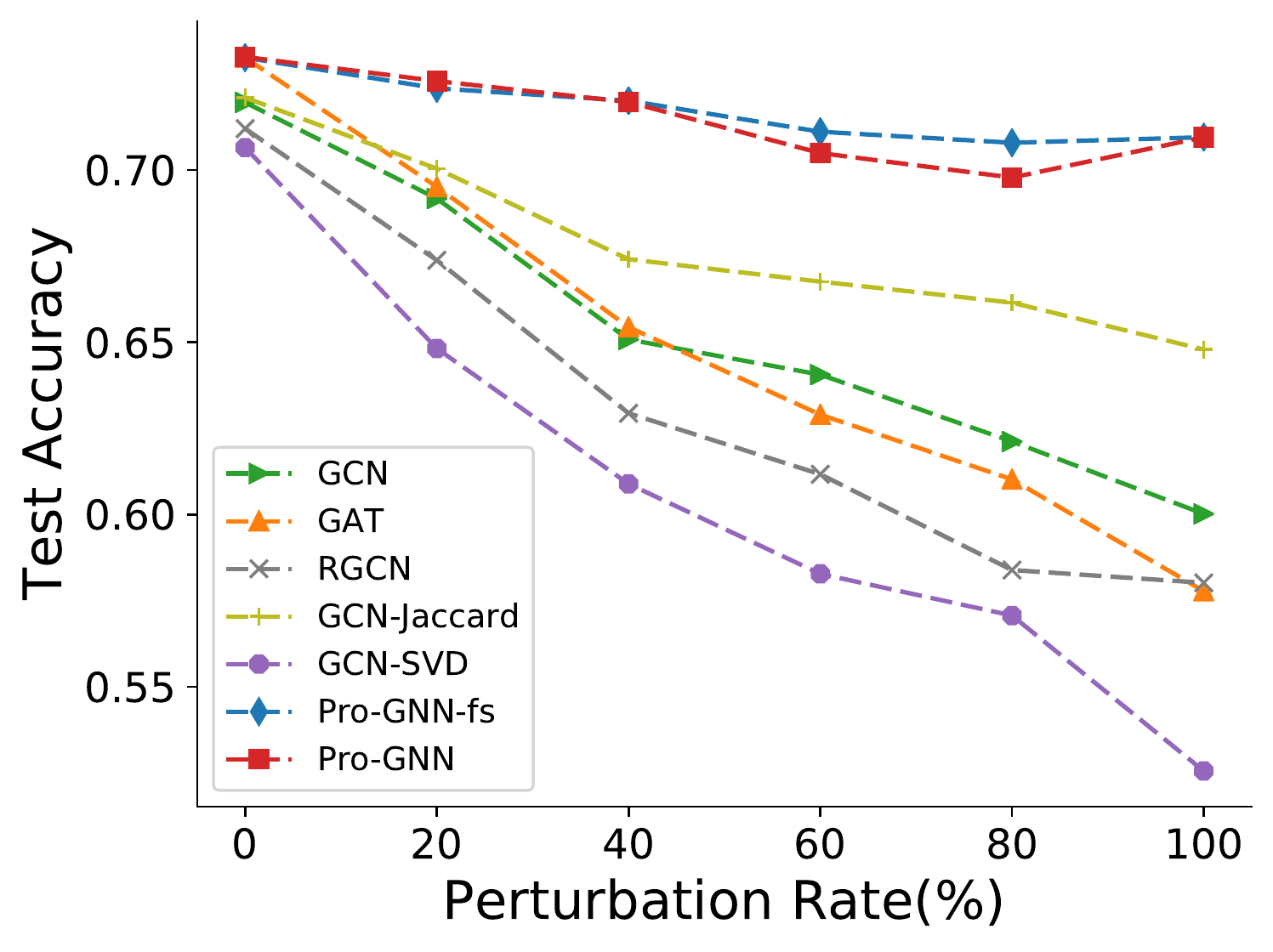} }}%
   \subfloat[Polblogs]{{\includegraphics[width=0.25\linewidth]{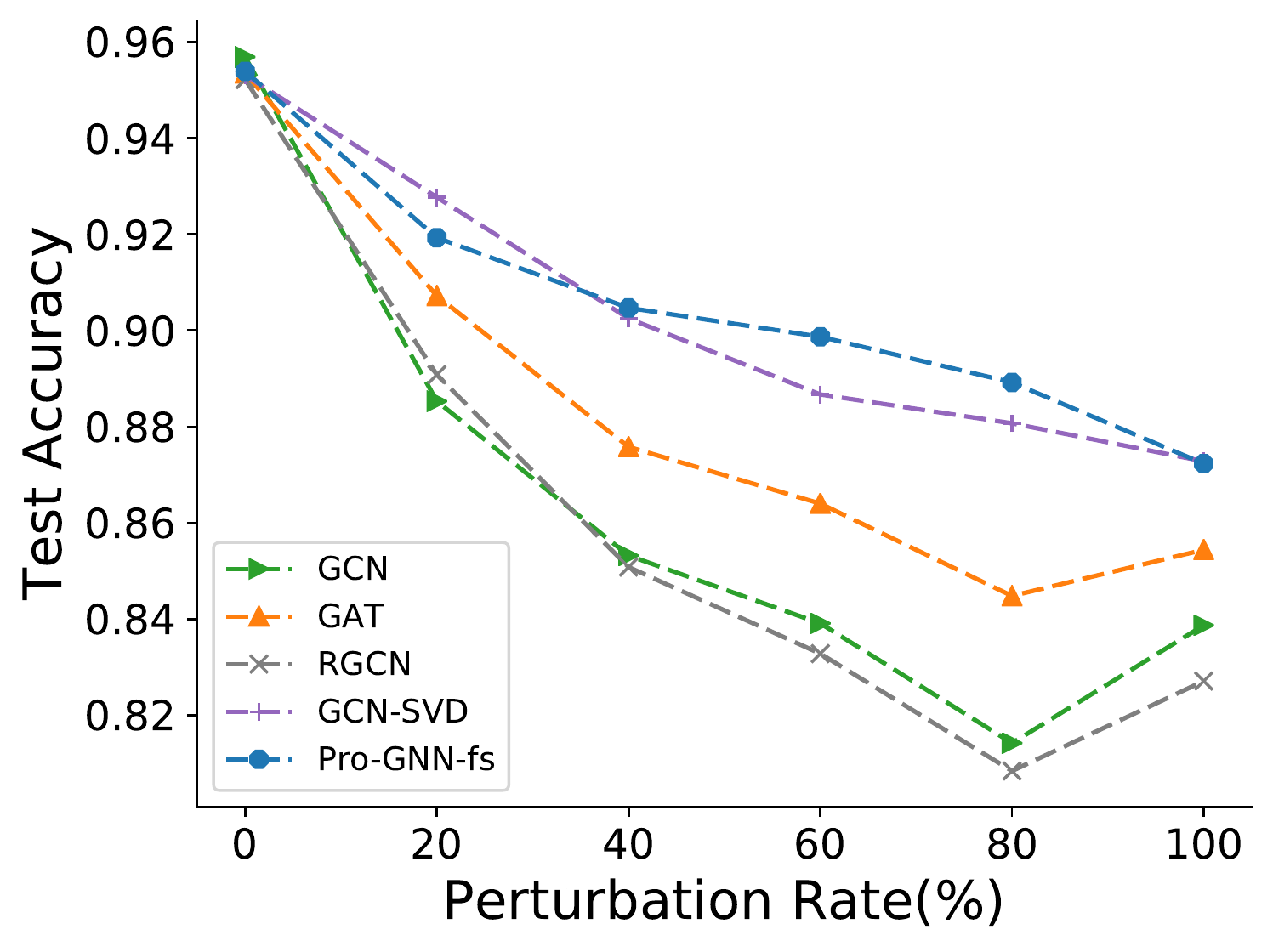} }}%
    \subfloat[Pubmed]{{\includegraphics[width=0.25\linewidth]{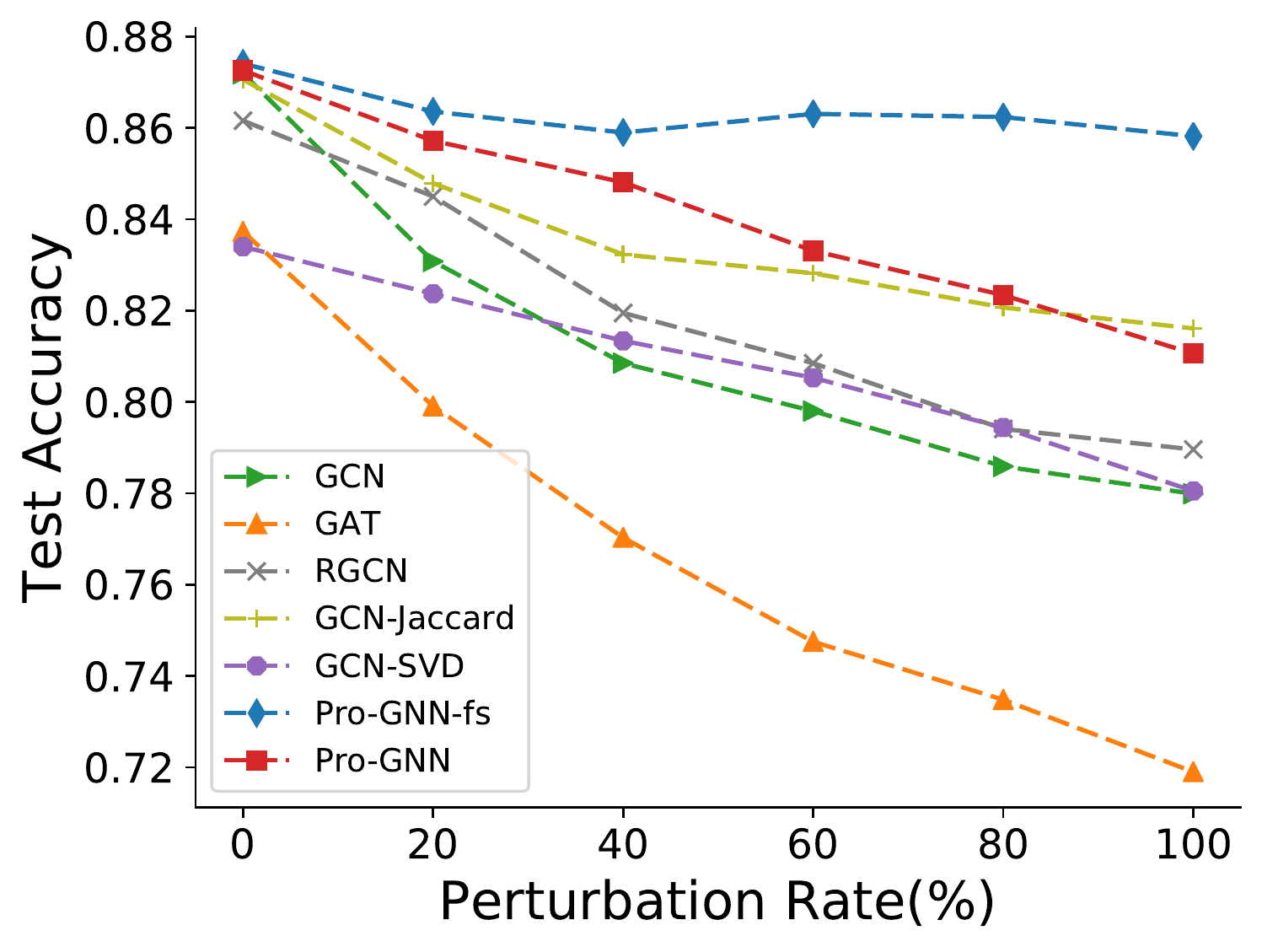} }}%
    \vskip -1.2em
    \caption{Results of different models under random attack}%
    \label{fig:random}%
    \vskip -1em
\end{figure*}

We first use the attack method to poison the graph. We then train Pro-GNN and baselines on the poisoned graph and evaluate the node classification performance achieved by these methods.   

\subsubsection{Against Non-targeted Adversarial Attacks}
\label{sec:non-target}
We first evaluate the node classification accuracy of different methods against non-targeted adversarial attack. Specifically, we adopt \metattack and keep all the default parameter settings in the authors' original implementation. The \metattack has several variants. For Cora, Citeseer and Polblogs datasets, we apply Meta-Self since it is the most destructive attack variant; while for Pubmed, the approximate version of Meta-Self, A-Meta-Self is applied to save memory and time. We vary the perturbation rate, i.e., the ratio of changed edges, from $0$ to $25\%$ with a step of $5\%$. As mentioned before, all the experiments are conducted 10 times and we report the average accuracy with standard deviation in Table~\ref{table:metattack}. The best performance is highlighted in bold. From the table, we make the following observations:

\begin{itemize}[leftmargin=*]
    \item 
Our method consistently outperforms other methods under different perturbation rates. For instance, on Polblogs dataset our model improves GCN over 20\% at 5\% perturbation rate. Even under large perturbation, our method outperforms other baselines by a larger margin. Specifically, under the $25\%$ perturbation rate on the three datasets, vanilla GCN performs very poorly and our model improves GCN by 22\%, 12\% and 14\%, respectively. 
\item
 Although \SVDpaper also employs SVD to get low-rank approximation of the graph, the performance of \SVDpaper drops rapidly. This is because \SVDpaper is designed for targeted attack,  it cannot adapt well to the non-targeted adversarial attack. Similarly, GCN-Jaccard does not perform as well as Pro-GNN under different perturbation rates. This is because simply preprocessing the perturbed graph once cannot recover the complex intrinsic graph structure from the carefully-crafted adversarial noises.  On the contrary, simultaneously updating the graph structure and GNN parameters with the low rank, sparsity and feature smoothness constraints helps recover better graph structure and learn robust GNN parameters.
\item Pro-GNN achieves higher accuracy than Pro-GNN-fs especially when the perturbation rate is large, which demonstrates the effectiveness of feature smoothing in removing adversarial edges.
\end{itemize}

\subsubsection{Against Targeted Adversarial Attack}
\label{sec:target}
In this experiment, \nettack is adopted as the targeted-attack method and we use the default parameter settings in the authors' original implementation. Following \cite{rgcn}, we vary the number of perturbations made on every targeted node from $1$ to $5$ with a step size of $1$. The nodes in test set with degree larger than $10$ are set as target nodes. For Pubmed dataset, we only sample 10\% of them to reduce the running time of \nettack while in other datasets we use all the target nodes. The node classification accuracy on target nodes is shown in Figure~\ref{fig:nettack}.
From the figure, we can observe that when the number of perturbation increases, the performance of our method is better than other methods on the attacked target nodes in most cases. For instance, on Citeseer dataset at 5 perturbation per targeted node, our model improves vanilla GCN by 23\% and outperforms other defense methods by 11\%. It demonstrates that our method can also resist the targeted adversarial attack.

\subsubsection{Against Random Attack}
\label{sec:random}
In this subsection, we evaluate how Pro-GNN behaves under different ratios of random noises from $0\%$ to $100\%$ with a step size of $20\%$. The results are reported in Figure~\ref{fig:random}. The figure shows that Pro-GNN consistently outperforms all other baselines and successfully resists random attack. Together with observations from Sections~\ref{sec:non-target} and~\ref{sec:target}, we can conclude that Pro-GNN is able to defend various types of adversarial attacks. This is a desired property in practice since attackers can adopt any kinds of attacks to fool the system.

\subsection{Importance of Graph Structure Learning}
In the previous subsection, we have demonstrated the effectiveness of the proposed framework. In this section, we aim to understand the graph we learned and answer the second question. 

\begin{figure}[htb]%
    \vskip -1.5em
    \centering
    \subfloat[Pubmed]{{\includegraphics[width=0.5\linewidth]{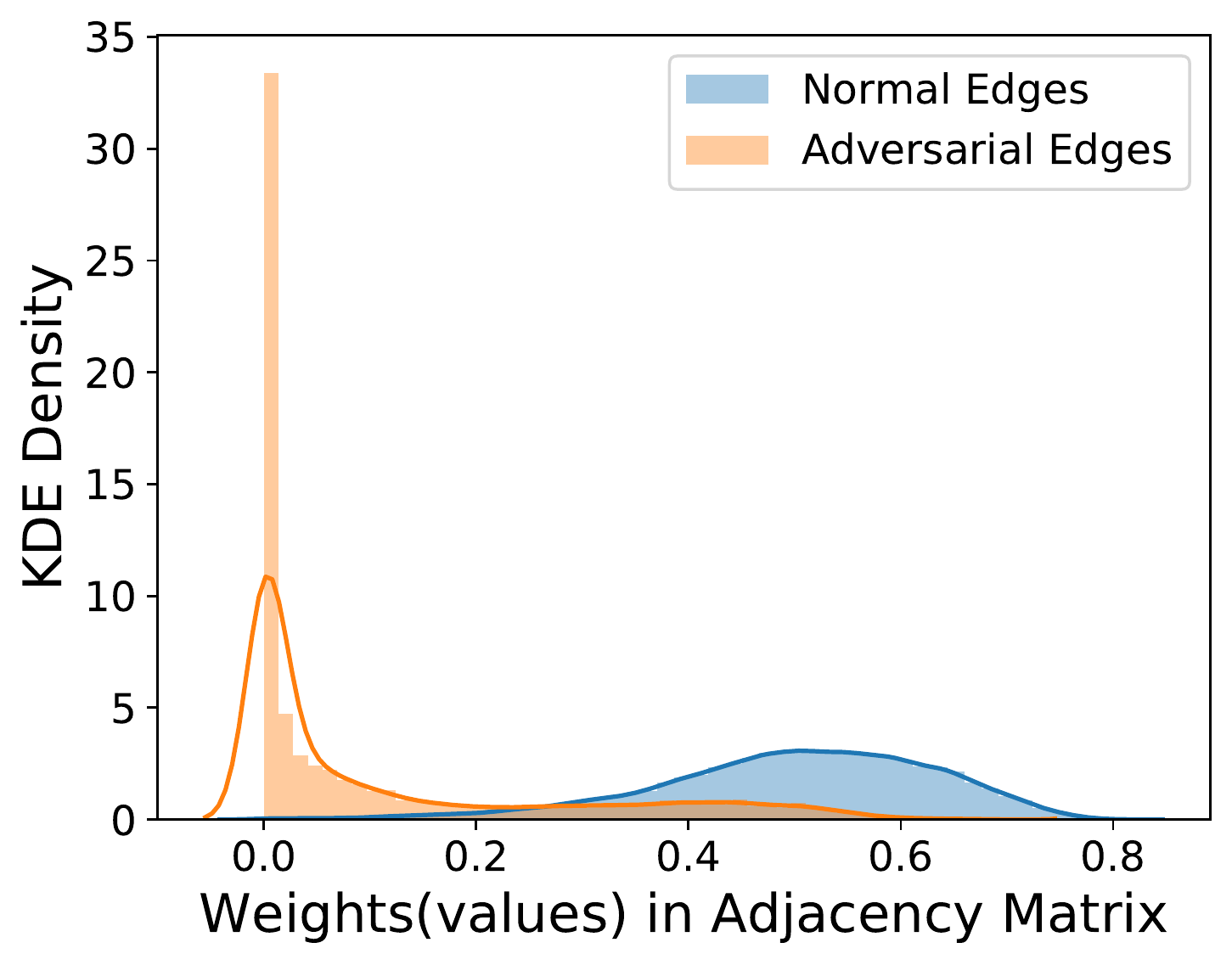} }}%
    \subfloat[Polblogs]{{\includegraphics[width=0.5\linewidth]{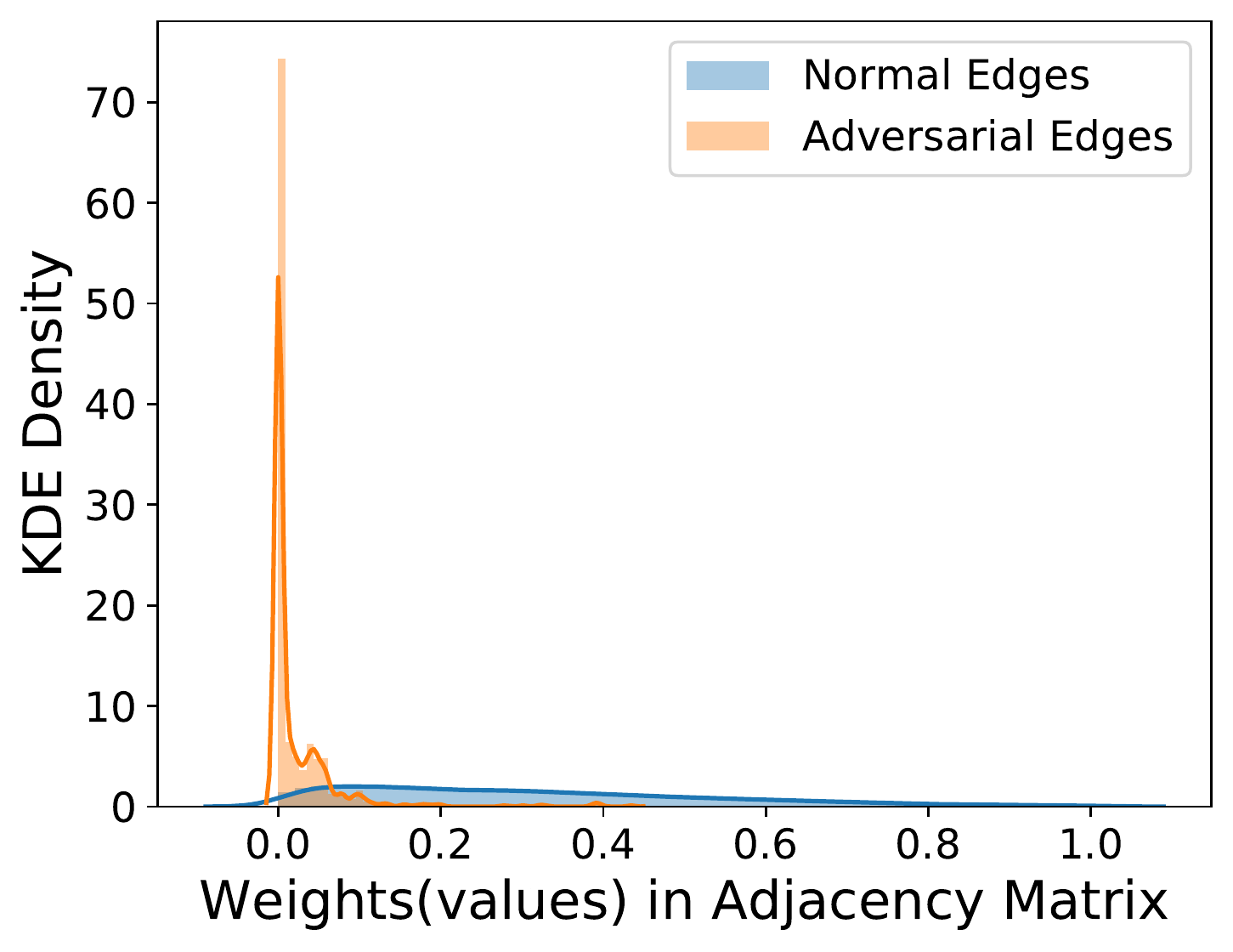} }}%
    \vskip -1.2em
    \caption{Weight density distributions of normal and adversarial edges on the learned graph.}%
    \label{fig:weights}%
    \vskip -1.2em
\end{figure}

\subsubsection{Normal Edges Against Adversarial Edges}
Based on the fact that adversary tends to add edges over delete edges~\cite{preprocess, mettack}, if the model tends to learn a clean graph structure, the impact of the adversarial edges should be mitigated from the poisoned graph. Thus, we investigate the weights of normal and adversarial edges in the learned adjacency matrix $\mathbf{S}$. We visualize the weight density distribution of normal and perturbed edges of $\mathbf{S}$ in Figure~\ref{fig:weights}. Due to the limit of space,  we only show results on Pubmed and Polblogs under \textit{metattack}. As we can see in the figure, in both datasets, the weights of adversarial edges are much smaller than those of normal edges, which shows that Pro-GNN can alleviate the effect of adversarial edges and thus learn robust GNN parameters.


\subsubsection{Performance on Heavily Poisoned Graph}
\label{sec:nograph}
In this subsection, we study the performance when the graph is heavily poisoned. In particular, we poison the graph with 25\% perturbation by \metattack. If a graph is heavily poisoned, the performance of GCN will degrade a lot. One straightforward solution is to remove the poisoned graph structure. Specifically, when removing the graph structure, the adjacency matrix will be all zeros and GCN normalizes the zero matrix into identity matrix and then makes prediction totally by node features. Under this circumstance, GCN actually becomes a feed-forward neural network. We denote it as GCN-NoGraph. We report the performance of GCN, GCN-NoGraph and Pro-GNN when the graph is heavily poisoned in Table~\ref{table:nograph}. 

From the table, we first observe that when the graph structure is heavily poisoned, by removing the graph structure, GCN-NoGraph outperforms GCN. This observation suggests the necessity to defend poisoning attacks on graphs because the poisoned graph structure are useless or even hurt the prediction performance. We also note that Pro-GNN obtains much better results  than GCN-NoGraph. This observation suggests that Pro-GNN can learn useful graph structural information even when the graph is heavily poisoned. 

\begin{table}[t]
\caption{Node classification accuracy given the graph under 25\% perturbation by \textit{metattack}.}
\label{table:nograph}
\vskip -1em
\small
\begin{tabular}{c|cccc}
\toprule
         & GCN & GCN-NoGraph       & \modelname           \\ \midrule 
Cora     & 47.53$\pm$1.96 & 62.12$\pm$1.55 & \textbf{69.72$\pm$1.69} \\ 
Citeseer & 56.94$\pm$2.09 & 63.75$\pm$3.23 & \textbf{68.95$\pm$2.78} \\ 
Polblogs & 49.23$\pm$1.36 & 51.79$\pm$0.62 & \textbf{63.18$\pm$4.40} \\ 
Pubmed  & 75.50$\pm$0.17 & 84.14$\pm$0.11 & \textbf{86.86$\pm$0.19} \\ \bottomrule
\end{tabular}
\vskip -1.5em
\end{table}


\subsection{Ablation Study}
\label{sec:ablation}
To get a better understanding of how different components help our model defend against adversarial attacks, we conduct ablation studies and answer the third question in this subsection. 

\subsubsection{Regularizers} \label{sec:regularizers}
\begin{figure}[tb]%
\vskip -1em
     \centering
     \subfloat[Cora]{{\includegraphics[width=0.5\linewidth]{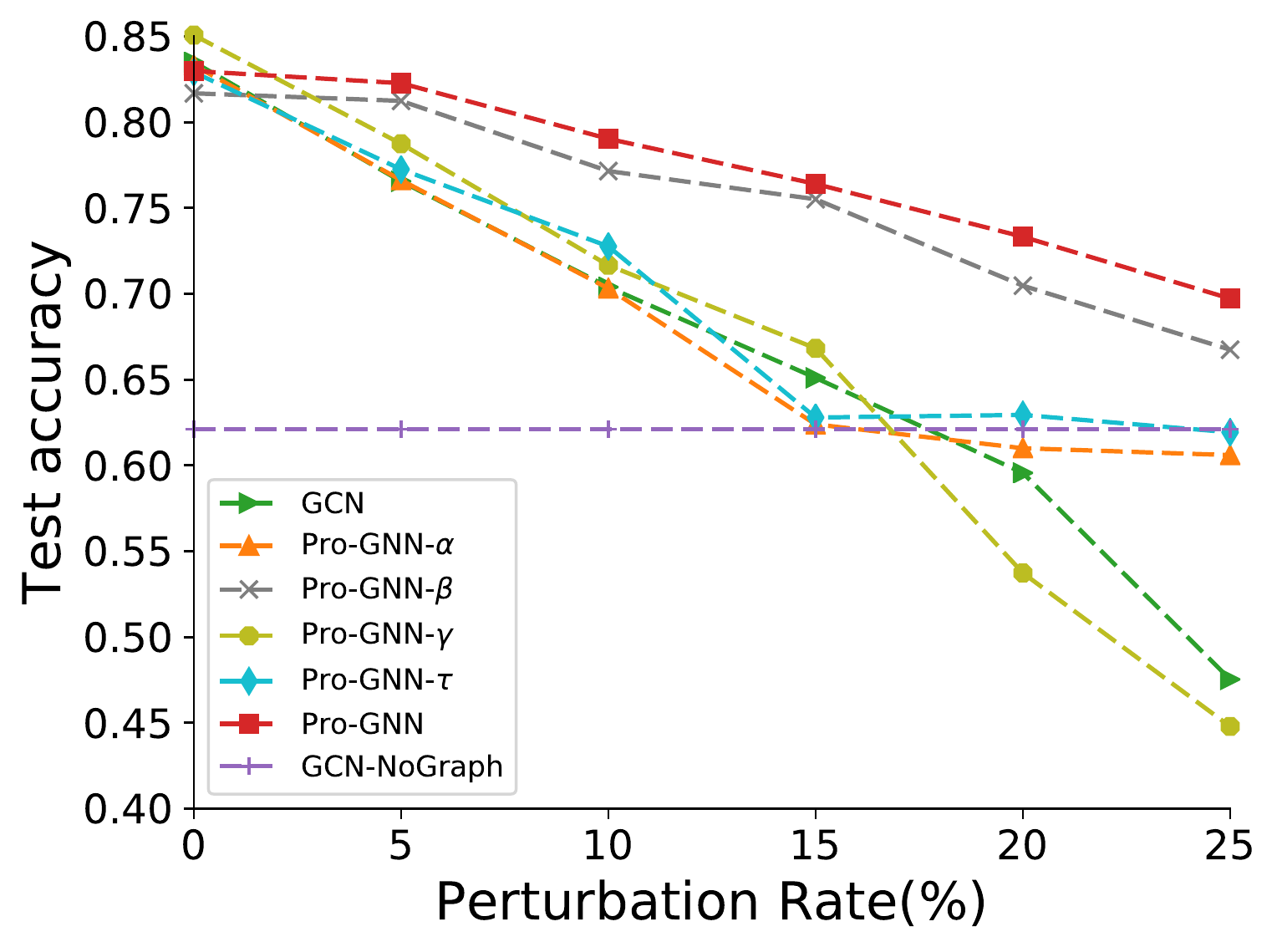} }}%
     \subfloat[Citeseer]{{\includegraphics[width=0.5\linewidth]{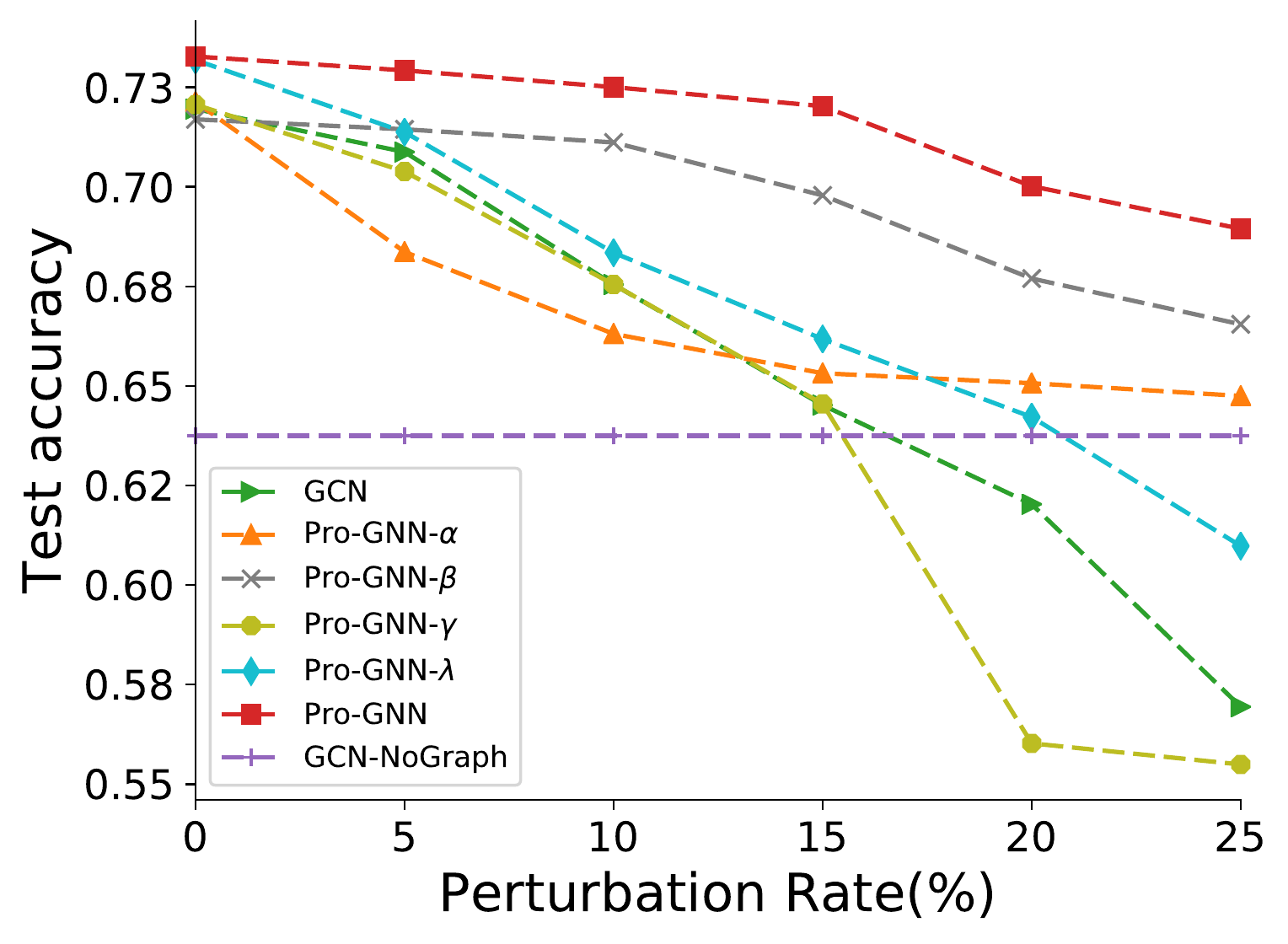} }}%
     \qquad
     \vskip -1.2em
    \caption{Classification performance of Pro-GNN variants.}
    \vskip -1em
\label{fig:ablation}%
\end{figure}

There are four key predefined parameters, i.e., ${\alpha}$, ${\beta}$, ${\gamma}$ and ${\lambda}$, which control the contributions for sparsity, low rank, GNN loss and feature smoothness, respectively.  To understand the impact of each component, we vary the values of one parameter and set other parameters to zero, and then check how the performance changes. Correspondingly, four model variants are created: Pro-GNN-${\alpha}$, Pro-GNN-${\beta}$, Pro-GNN-${\gamma}$ and Pro-GNN-${\lambda}$. For example, Pro-GNN-${\alpha}$ denotes that we vary the values of ${\alpha}$ while setting  $\beta$, ${\gamma}$ and ${\lambda}$ to zero. We only report results on Cora and Citeseer, since similar patterns are observed in other cases, shown in Figure~\ref{fig:ablation}.

From the figure we can see Pro-GNN-${\alpha}$ does not boost the model's performance too much with small perturbations. But when the perturbation becomes large, Pro-GNN-${\alpha}$ outperforms vanilla GCN because it can learn a graph structure better than a heavily poisoned adjacency graph as shown in Section~\ref{sec:nograph}. Also, Pro-GNN-${\beta}$ and Pro-GNN-${\lambda}$ perform much better than vanilla GCN. It is worth noting that, Pro-GNN-${\beta}$ outperforms all other variants except Pro-GNN, indicating that nuclear norm is of great significance in reducing the impact of adversarial attacks. It is in line with our observation that adversarial attacks increase the rank of the graph and enlarge the singular values. Another observation from the figure is that, Pro-GNN-${\gamma}$ works better under small perturbation and when the perturbation rate increases, its performance degrades. From the above observations, different components play different roles in defending adversarial attacks. By incorporating these components, Pro-GNN can explore the graph properties and thus consistently outperform state-of-the-art baselines.

\subsubsection{Two-Stage vs One-Stage} \label{sec:two_stage_one_stage}
\begin{table*}[t]
\caption{Classification performance of Pro-GNN-two and Pro-GNN on Cora dataset}
\vskip -1em
\label{table:two_stage}
\begin{tabular}{@{}lllllll@{}}
\toprule
Ptb Rate (\%) & 0                       & 5                       & 10                      & 15                      & 20                      & 25                      \\ \midrule
Pro-GNN-two   & 73.31$\pm$0.71          & 73.70$\pm$1.02          & 73.69$\pm$0.81          & 75.38$\pm$1.10          & 73.22$\pm$1.08          & \textbf{70.57$\pm$0.61} \\
Pro-GNN       & \textbf{82.98$\pm$0.23} & \textbf{82.27$\pm$0.45} & \textbf{79.03$\pm$0.59} & \textbf{76.40$\pm$1.27} & \textbf{73.32$\pm$1.56} & 69.72$\pm$1.69          \\ \bottomrule
\end{tabular}
\end{table*}
To study the contribution of jointly learning structure and GNN parameters, we conduct experiments with the variant Pro-GNN-two under \metattack. Pro-GNN-two is the two stage variant of Pro-GNN where we first obtain the clean graph and then train a GNN model based on it. We only show the results on Cora in Table~\ref{table:two_stage} due to the page limitation. We can observe from the results that although Pro-GNN-two can achieve good performance under large perturbation, it fails to defend the attacks when the perturbation rate is relatively low. The results demonstrate that jointly learning structure and GNN parameters can actually help defend attacks.

\subsection{Parameter Analysis}
In this subsection, we explore the sensitivity of hyper-parameters $\alpha, \beta, \gamma$ and $\lambda$ for Pro-GNN. In the experiments, we alter the value of $\alpha, \beta, \gamma$ and $\lambda$ to see how they affect the performance of our model. 
More specifically, we vary $\alpha$ from $0.00025$ to $0.064$ in a log scale of base 2, $\beta$ from $0$ to $5$, $\gamma$ from $0.0625$ to $16$ in a log scale of base 2 and $\lambda$ from $1.25$ to $320$ in a log scale of base 2. We only report the results on Cora dataset with the perturbation rate of $10\%$ by \metattack since similar observations are made in other settings. 

\begin{figure}[tb]%
\vskip -1em
    \centering
    \subfloat{{\includegraphics[width=0.5\linewidth]{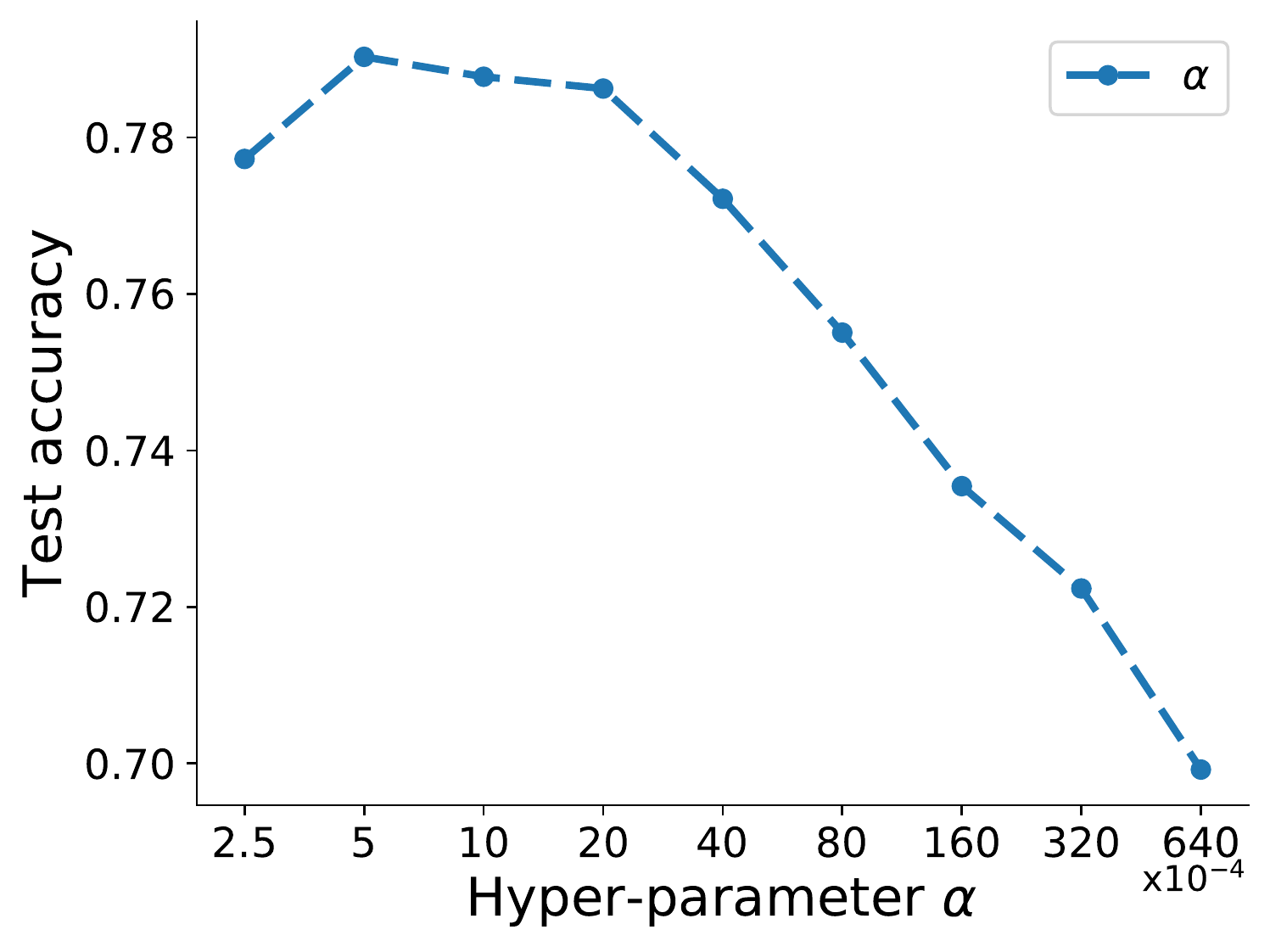} }}%
    \subfloat{{\includegraphics[width=0.5\linewidth]{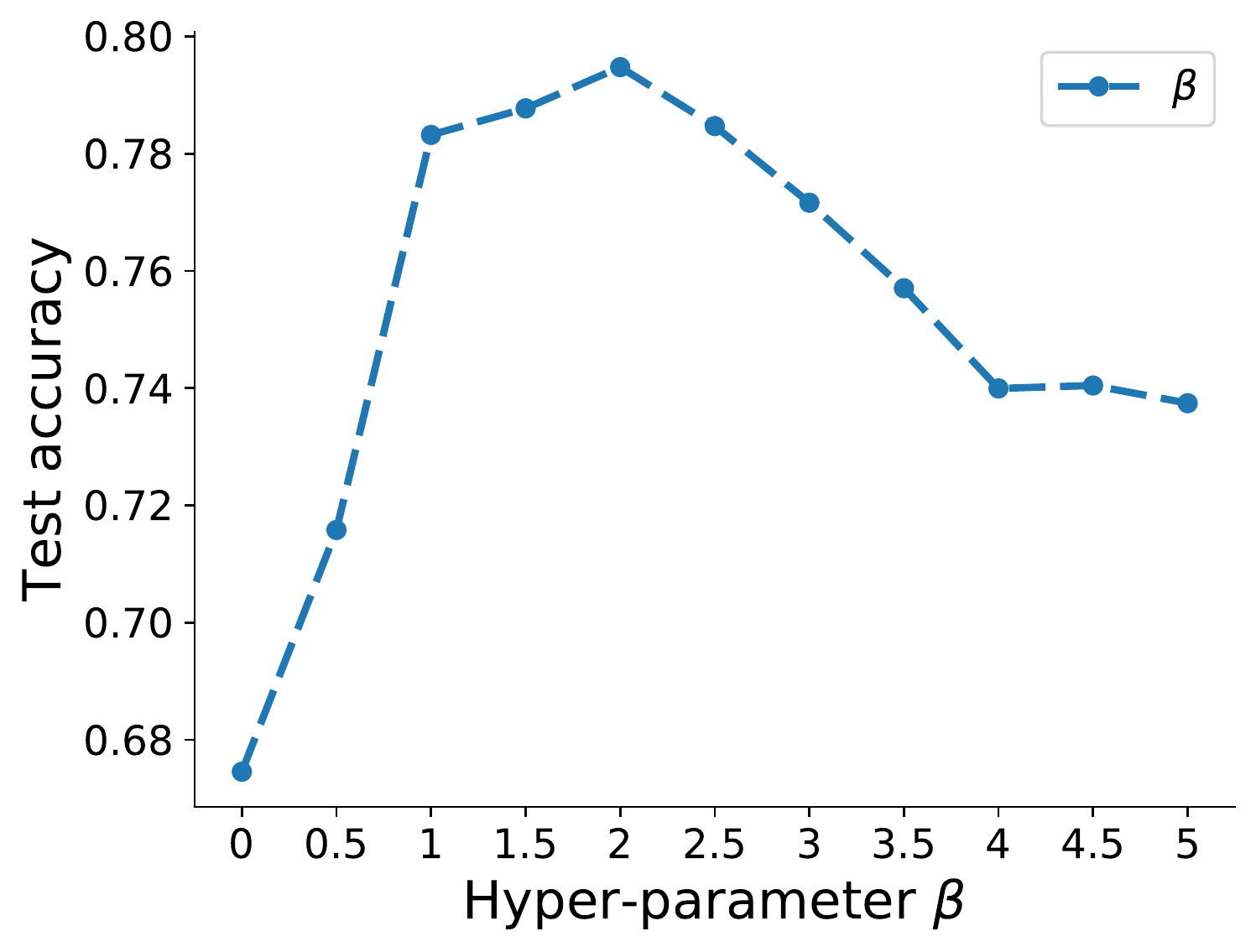} }}%
    \vskip -1em
    \subfloat{{\includegraphics[width=0.5\linewidth]{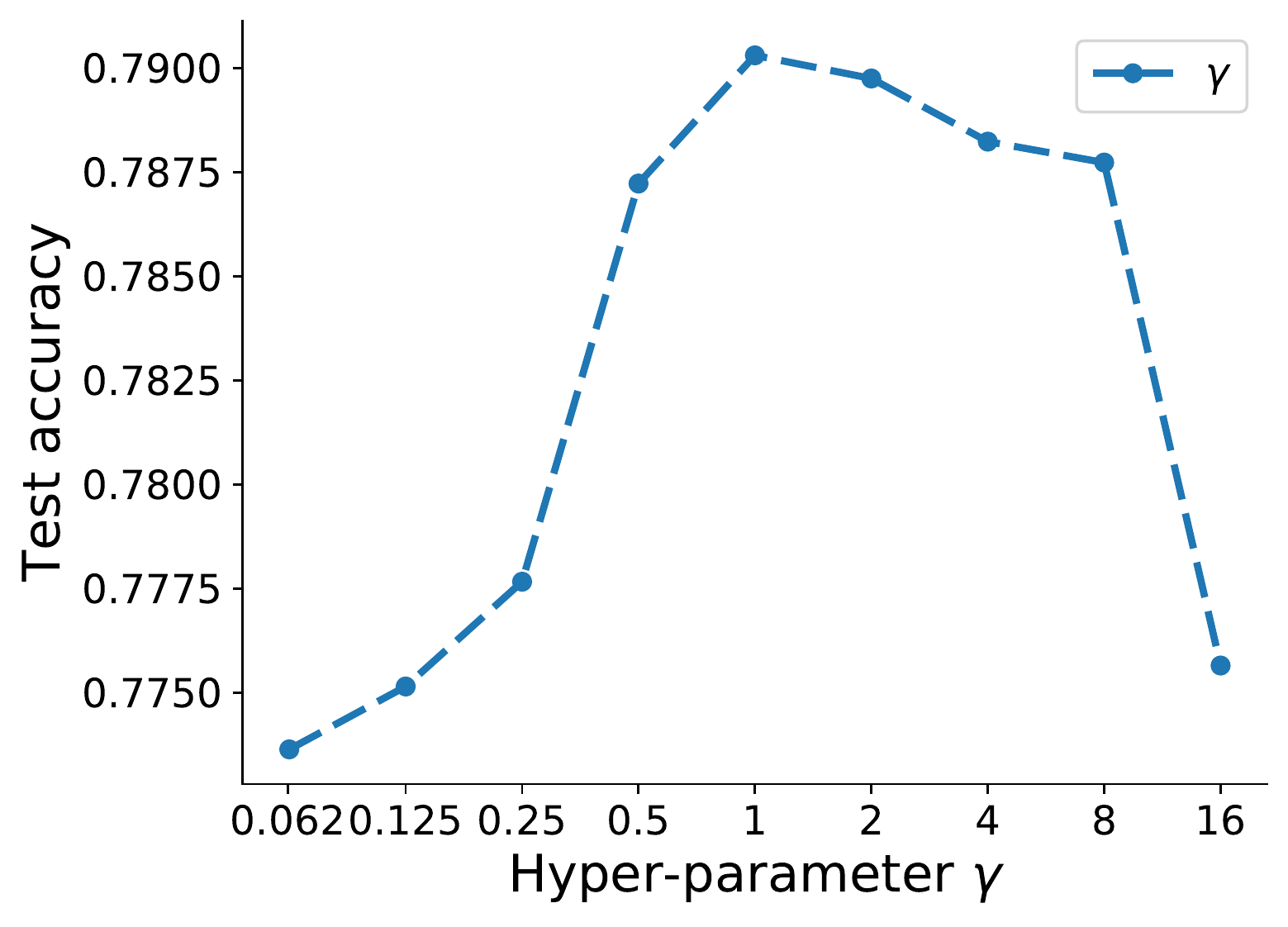} }}%
    \subfloat{{\includegraphics[width=0.5\linewidth]{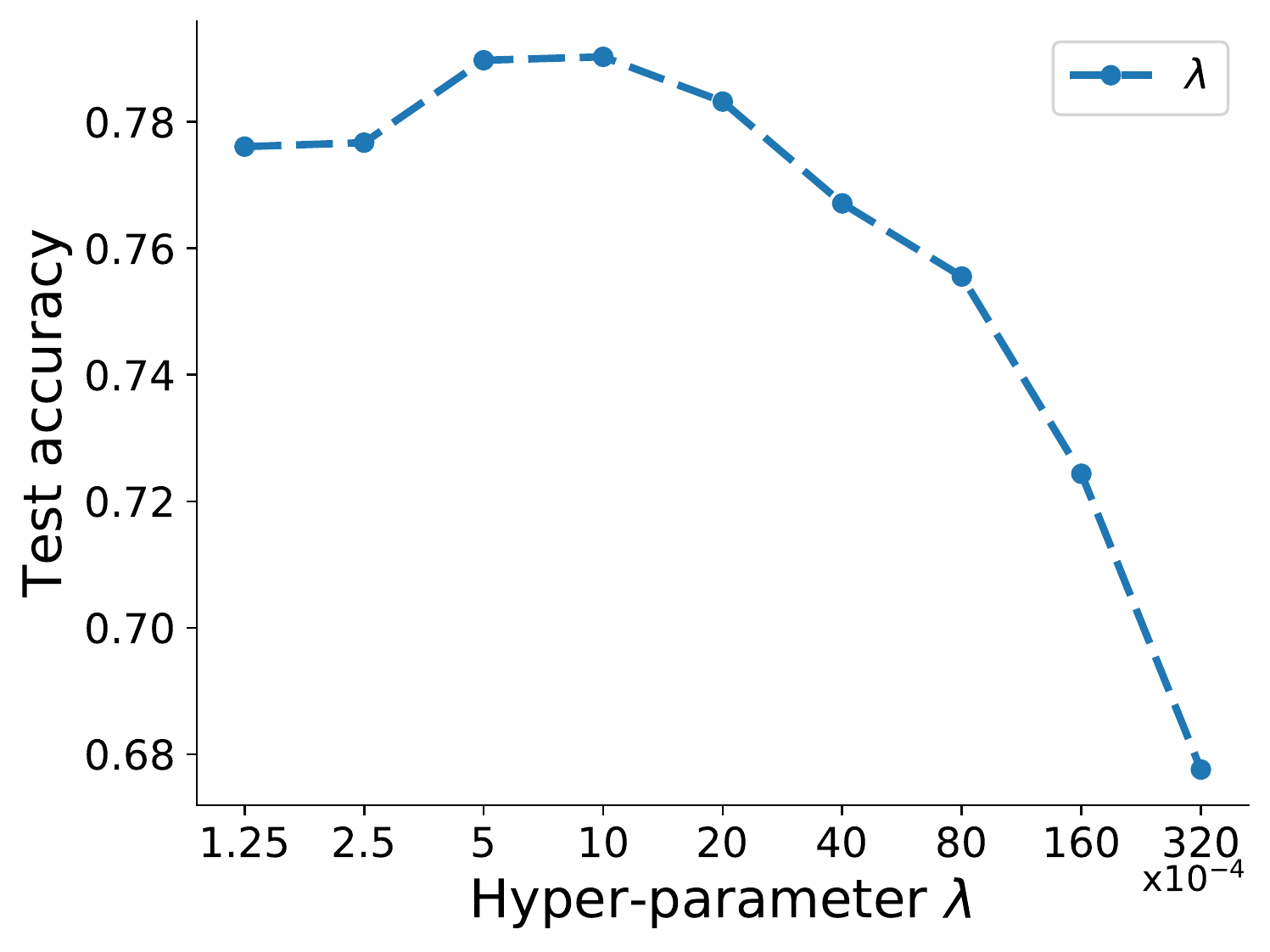} }}%
    \vskip -1em
    \caption{Results of parameter analysis on Cora dataset}%
    \label{fig:params}%
\vskip -1em
\end{figure}

The performance change of Pro-GNN is illustrated in Figure~\ref{fig:params}. As we can see, the accuracy of Pro-GNN can be boosted when choosing appropriate values for all the hyper-parameters. Different from $\gamma$,  appropriate values of $\alpha$ and $\lambda$ can boost the performance but large values will greatly hurt the performance. This is because focusing on sparsity and feature smoothness will result in inaccurate estimation on the graph structure.  For example, if we set $\alpha$ and $\lambda$ to $+\infty$, we will get a trivial solution of the new adjacency matrix, i.e, ${\bf S} = 0$.  It is worth noting that, appropriate value of $\beta$ can greatly increase the model's performance (more than 10\%) compared with the variant without $\beta$, while too large or too small value of $\beta$ will hurt the performance. This is also consistent with our observation in Section \ref{sec:regularizers} that the low rank property plays an important role in defending adversarial attacks.

\section{Conclusion}

Graph neural networks can be easily fooled by graph adversarial attacks. To defend against different types of graph adversarial attacks, we introduced a novel defense approach \modelname that learns the graph structure and the GNN parameters simultaneously. Our experiments show that our model consistently outperforms state-of-the-art baselines and improves the overall robustness under various adversarial attacks. In the future, we aim to explore more properties to further improve the robustness of GNNs.

\section{Acknowledgements}
This research is supported by the National Science Foundation (NSF) under grant numbers IIS1907704, IIS1928278, IIS1714741, IIS1715940, IIS1845081, IIS1909702 and CNS1815636.

\bibliographystyle{ACM-Reference-Format}
\bibliography{main}


\end{document}